\pdfoutput=1

\documentclass[11pt]{article}
\usepackage[T1]{fontenc}
\usepackage[utf8]{inputenc}
\usepackage{listings}
\usepackage{xcolor}
\usepackage{longtable}
\usepackage{inconsolata}
\usepackage{microtype}

\usepackage[final]{acl}

\usepackage{times}
\usepackage{latexsym}
\usepackage{float}
\restylefloat{table}
\usepackage{fvextra}
\usepackage{makecell}

\usepackage[T1]{fontenc} 
\usepackage{kotex}

\usepackage[utf8]{inputenc}

\usepackage{color-edits}
\addauthor{gn}{magenta}
\addauthor{byl}{purple}

\usepackage{microtype}
\usepackage{graphicx}
\graphicspath{{image/}}
\usepackage{pdfpages}
\usepackage{url}
\usepackage{hyperref}
\usepackage{xcolor}
\usepackage{epsfig}
\usepackage{adjustbox}
\usepackage{amsfonts}
\usepackage{amsmath}
\usepackage{amssymb}
\usepackage{booktabs} 
\usepackage{comment}
\usepackage{multirow}
\usepackage{caption}
\usepackage{subcaption}
\usepackage{textcomp}
\usepackage{comment}
\usepackage{relsize}
\usepackage{stmaryrd}
\usepackage{bbm}
\usepackage{rotating}
\usepackage{arydshln}
\usepackage{pifont}
\usepackage{xcolor}
\usepackage{colortbl}
\usepackage{tabularray}
\usepackage{graphicx}
\usepackage{pbox}
\usepackage[most]{tcolorbox}
\usepackage{colortbl}
\usepackage{enumitem}

\usepackage[htt]{hyphenat}
\usepackage{xurl}
\usepackage{hyperref}
\newtcolorbox{mybox}[2][]{
    colback=white,
    colframe=green!45,
    fonttitle=\bfseries,
    coltitle=black,
    sharp corners,
    title=#2,
    #1
}
\usepackage[most]{tcolorbox}
\usepackage{tikz}
\usepackage{enumitem}
\usepackage{xcolor}
\usepackage{wasysym}

\definecolor{kbBlue}{HTML}{244A9B}
\definecolor{kbGreen}{HTML}{1F7A3A}
\definecolor{kbRed}{HTML}{A91E1E}
\definecolor{kbOrange}{HTML}{D96B00}
\definecolor{kbGray}{HTML}{F6F6F6}
\definecolor{kbState}{HTML}{4A4A4A}

\newcommand{\kbBadge}[2]{%
  \tikz[baseline=(X.base)]{
    \node[
      circle,
      draw=#1!70!black,
      fill=#1!10,
      text=#1!80!black,
      inner sep=1.15pt,
      font=\scriptsize\bfseries
    ] (X) {#2};
  }%
}

\newcommand{\gstep}[1]{%
  \tikz[baseline=(X.base)]{
    \node[
      circle,
      draw=kbGreen!70!black,
      text=kbGreen!70!black,
      inner sep=0.35pt,
      font=\tiny\bfseries
    ] (X) {#1};
  }%
}

\newcommand{\rstep}[1]{%
  \tikz[baseline=(X.base)]{
    \node[
      circle,
      draw=kbRed!75!black,
      text=kbRed!75!black,
      inner sep=0.35pt,
      font=\tiny\bfseries
    ] (X) {#1};
  }%
}

\newenvironment{trajblock}[2]{%
\begin{tcolorbox}[
  enhanced,
  colback=#1!4,
  colframe=#1!65!black,
  boxrule=0.38pt,
  arc=1.4pt,
  left=3.5pt,
  right=3.5pt,
  top=2.2pt,
  bottom=2.2pt,
  before skip=1.6pt,
  after skip=2.2pt
]
\noindent\textcolor{#1!75!black}{\textbf{#2}}\par\vspace{0.5pt}
}{%
\end{tcolorbox}
}

\newcommand{\searchquery}[1]{%
\vspace{0.5pt}
\begin{tcolorbox}[
  enhanced,
  colback=gray!7,
  colframe=gray!38,
  boxrule=0.25pt,
  arc=0.8pt,
  left=2.5pt,
  right=2.5pt,
  top=1.2pt,
  bottom=1.2pt,
  before skip=0.8pt,
  after skip=0.8pt
]
{\ttfamily\fontsize{5.8}{6.4}\selectfont\raggedright #1\par}
\end{tcolorbox}
}

\newenvironment{failurecase}[3]{%
\begin{minipage}[t]{0.492\textwidth}
\begin{tcolorbox}[
  enhanced,
  colback=white,
  colframe=black!62,
  boxrule=0.55pt,
  arc=2pt,
  left=4pt,
  right=4pt,
  top=3.5pt,
  bottom=3.5pt
]
\centering
\textbf{(#1) \texttt{#2}. #3}
\vspace{1.5pt}
\raggedright
}{%
\end{tcolorbox}
\end{minipage}
}

\newcommand{\eng}[1]{\textcolor{black!38}{\tiny\textit{(#1)}}}

\title{\textsc{K-BrowseComp}:\\ A Web Browsing Agent Benchmark Grounded in Korean Contexts}

\author{Nahyun Lee$^{1,4}$ \quad Dongkeun Yoon$^{2}$ \quad \textbf{Guijin Son}$^{3,4}$ \quad \textbf{Geewook Kim}$^{2,5}$ \quad \textbf{Dayoon Ko}$^{3}$ \quad  \\ \textbf{Jeonghun Park}$^{3}$ \quad \textbf{Haneul Yoo}$^{6}$ \quad
\textbf{Jaewon Cho}$^{2}$ \quad 
\textbf{Junghun Park}$^{3}$ \quad 
\textbf{Changyoon Lee}$^{2}$ \quad   \\ 
\textbf{Kyochul Jang}$^{3}$ \quad
\textbf{Jaeyeon Kim}$^{7}$ \quad
\textbf{Eunsu Kim}$^{7}$ \quad 
\textbf{Seungone Kim}$^{7}$\thanks{~~Corresponding author}\\ \\ Chung-Ang University$^{1}$ \qquad KAIST$^{2}$ \qquad Seoul National University$^{3}$\qquad OnelineAI$^{4}$ \\ \qquad NAVER Cloud AI$^{5}$ \qquad New York University$^{6}$ \qquad Carnegie Mellon University$^{7}$\\
\texttt{
naa012@cau.ac.kr \qquad seungone@cmu.edu}}

\begin{document}
\maketitle
\newcommand{\mcal}[1]{{\cal{#1}}}
\newcommand{\calA}{\mbox{${\cal A}$}}
\newcommand{\calB}{\mbox{${\cal B}$}}
\newcommand{\calC}{\mbox{${\cal C}$}}
\newcommand{\calD}{\mbox{${\cal D}$}}
\newcommand{\calE}{\mbox{${\cal E}$}}
\newcommand{\calF}{\mbox{${\cal F}$}}
\newcommand{\calG}{\mbox{${\cal G}$}}
\newcommand{\calH}{\mbox{${\cal H}$}}
\newcommand{\calI}{\mbox{${\cal I}$}}
\newcommand{\calJ}{\mbox{${\cal J}$}}
\newcommand{\calK}{\mbox{${\cal K}$}}
\newcommand{\calL}{\mbox{${\cal L}$}}
\newcommand{\calM}{\mbox{${\cal M}$}}
\newcommand{\calN}{\mbox{${\cal N}$}}
\newcommand{\calO}{\mbox{${\cal O}$}}
\newcommand{\calP}{\mbox{${\cal P}$}}
\newcommand{\calQ}{\mbox{${\cal Q}$}}
\newcommand{\calR}{\mbox{${\cal R}$}}
\newcommand{\calS}{\mbox{${\cal S}$}}
\newcommand{\calT}{\mbox{${\cal T}$}}
\newcommand{\calU}{\mbox{${\cal U}$}}
\newcommand{\calV}{\mbox{${\cal V}$}}
\newcommand{\calW}{\mbox{${\cal W}$}}
\newcommand{\calX}{\mbox{${\cal X}$}}
\newcommand{\calY}{\mbox{${\cal Y}$}}
\newcommand{\calZ}{\mbox{${\cal Z}$}}

\newcommand*\concat{\mathbin{\|}}

\newcommand{\che}[1]{\textcolor{brown}{#1}}

\newcommand{\se}{{\it SE}}
\newcommand{\eg}{{\it e.g.}}
\newcommand{\ie}{{\it i.e.}}
\newcommand{\etal}{{\it et al.}}
\newcommand{\etc}{{\it etc}}
\newcommand{\yeoo}{\textcolor{blue}}
\newcommand{\argmin}{\mathop{\mathrm{argmin}}\limits}
\newcommand{\argmax}{\mathop{\mathrm{argmax}}\limits}
\newcommand{\gc}{\textcolor{green}{\ding{52}}}
\newcommand{\bt}{\textcolor{black}{\ding{115}}}
\newcommand{\rx}{\textcolor{red}{\ding{55}}}
\definecolor{lightblue}{RGB}{224,236,247}
\definecolor{deepblue}{RGB}{9,46,107}
\begin{abstract}
Frontier model evaluations are shifting from foundational capabilities
(e.g., instruction following and reasoning) toward compositional, agentic ones, but Korean agentic benchmarks remain scarce. We introduce \textsc{K-BrowseComp}, a web-browsing agent benchmark grounded in Korean contexts, consisting of 400 problems. The 300-problem \textsc{K-BrowseComp-Verified} subset is manually constructed and validated by native Korean speakers. On this subset, frontier LLMs, including GPT-5.5, DeepSeek-V4-Pro, and GLM-5.1, reach only 30.00--45.67\%, a substantial drop from BrowseComp, while Korean LLMs released through Korea's Proprietary AI Foundation Model program achieve at most 10.33\%. We further construct a 100-problem \textsc{Synthetic} split using hard few-shot exemplars and failure-mode-targeted generation to exploit the asymmetry between solving and creating web browsing problems.
On the adversarially filtered synthetic diagnostic split, the strongest model reaches only 26.00\%, and we report this split separately as a targeted stress test.
We publicly release our data and code.\footnote{\url{https://github.com/prometheus-eval/K-BrowseComp}}

\end{abstract}

\section{Introduction}

While leading frontier models from the US and China are shifting their evaluation focus from foundational capabilities (\textit{e.g.}, instruction following, reasoning, and tool calling) toward compositional agentic evaluation~\citep{singh2025openai,deepseekai2026deepseekv4}, the Korean AI community remains largely anchored to static benchmarks~\citep{sovereign2025,choi2026k,park2026solar,team2026hyperclova}. Korean agentic benchmarks are still virtually nonexistent, leaving the community without a standardized way to measure progress.
Developing such benchmarks is important for two reasons.

\begin{figure}[t!]
\centering
\includegraphics[width=0.99\linewidth]{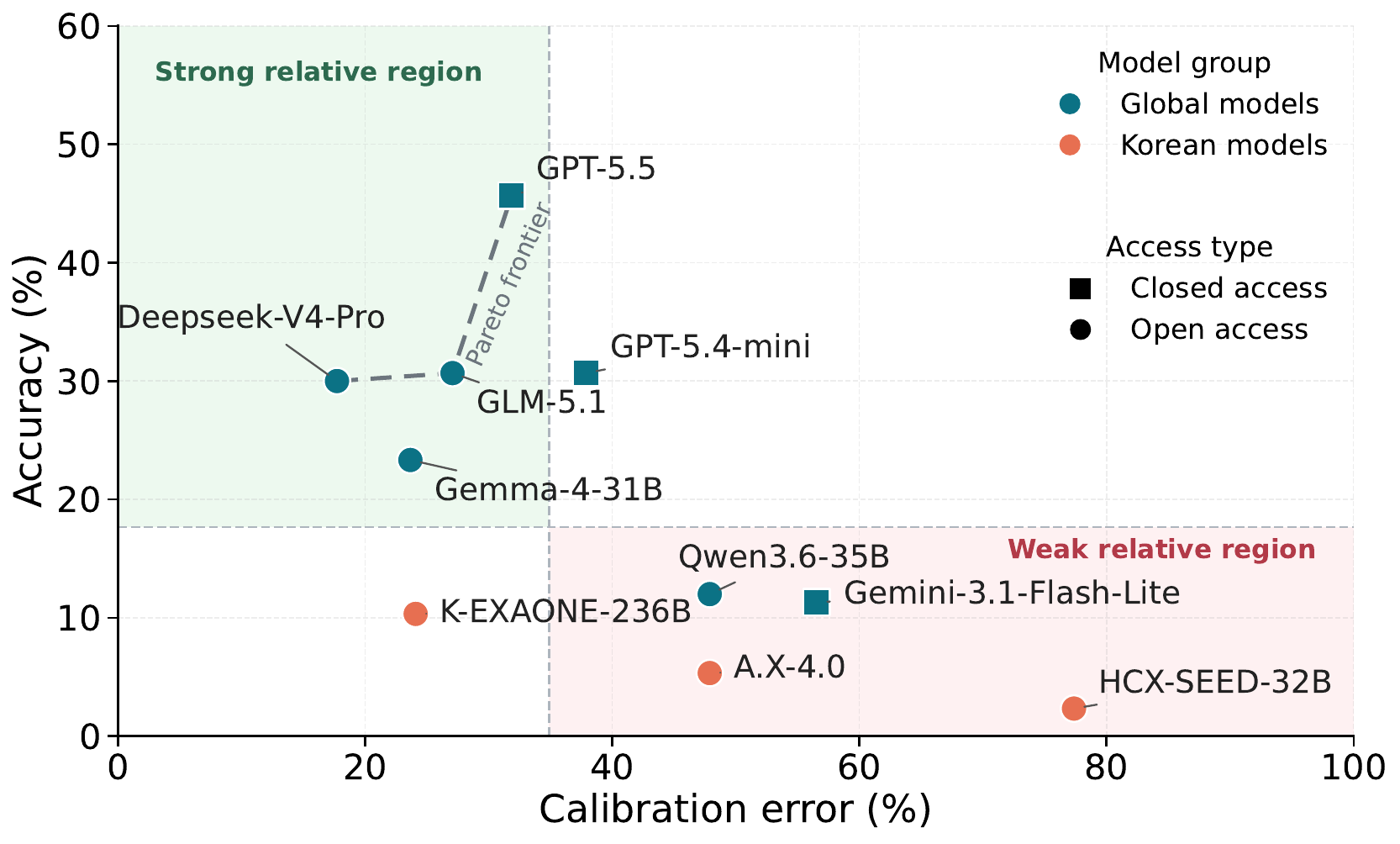}
    \caption{
\textbf{Accuracy and calibration error of evaluated models on \textsc{K-BrowseComp-Verified}.}
Higher accuracy and lower calibration error indicate better performance.
The shaded quadrants are defined by the median accuracy and calibration error across models. The dashed line marks the Pareto frontier.
}
    \label{fig:main_plot}
\vspace{-3mm}
\end{figure}

\begin{itemize}[leftmargin=*]
    \item For \textbf{Korean developers and users}, language usage and population size place Korea at a structural disadvantage relative to larger language communities, raising AI sovereignty concerns for queries requiring Korean local and cultural knowledge~\citep{kim2024click,son2025kmmlu}. 
    \item Second, Korean agentic benchmarks serve the \textbf{broader research community}. Frontier models increasingly saturate existing benchmarks, blurring the boundary between in-distribution and out-of-distribution evaluation~\citep{dong2024generalization,spiesberger2026soft}. Agentic benchmarks grounded in linguistically and culturally distinct contexts can therefore provide a principled testbed for broader generalization~\citep{romanou2025include,whitehouse2026menlo}.
\end{itemize}

Based on this motivation, we take a first step toward building Korean agentic benchmarks by proposing \textsc{K-BrowseComp}, a browsing agent benchmark grounded in Korean contexts.
We focus on browsing agents for two reasons.
First, browsing agents are uniquely dependent on local and cultural knowledge: their core function is to retrieve region-specific information from the web, making the gap between Korean and English-centric contexts especially consequential for end users.
Second, browsing agents are inherently compositional, jointly exercising instruction following, tool calling, and multi-turn interaction, and thus provide a comprehensive testbed for Korean agentic models.

\textsc{K-BrowseComp} consists of 400 problems in total, including the \textsc{K-BrowseComp-Verified} subset of 300 manually crafted problems, which are validated by native Korean speakers, and a 100-problem \textsc{Synthetic} diagnostic split. As shown in Figure~\ref{fig:main_plot} and Table~\ref{tab:main_results}, the strongest evaluated models achieve only 30.00--45.67\% on \textsc{K-BrowseComp-Verified}. 
Specifically, \texttt{GPT-5.5} and \texttt{DeepSeek-V4-Pro} obtain 45.67\% and 30.00\%, respectively, which is a substantial drop from their reported performance on the original BrowseComp~\citep{wei2025browsecomp} (84.4\% and 83.4\%, respectively).
Moreover, Korean LLMs released through the first round of government-funded ``Proprietary AI Foundation Model Project''~\citep{sovereign2025} achieve at most 10.33\%.
Our trajectory analysis shows that these gaps do not reduce to a single bottleneck.
Some models terminate search prematurely or fail to emit stable tool-call trajectories, while others retrieve relevant Korean web evidence but fail to preserve candidate sets, constraints, role bindings, or final-answer state across sources.
This indicates that \textsc{K-BrowseComp} can serve as a diagnostic target for the Korean ecosystem to develop browsing agents tailored to Korean users.


Beyond \textsc{K-BrowseComp-Verified} subset, the core technical contribution of this paper lies in our methodology for constructing the remaining 100 problems with LLMs. A defining property of browsing tasks is their \textit{information asymmetry}: solving a problem can be difficult, while verifying a candidate answer is comparatively easier once the relevant evidence path is known~\citep{wei2025browsecomp}. Motivated by this asymmetry, we examine an analogous question on the construction side: \textit{while solving such problems is hard, is it also hard to create them}? 

To this end, we employ a web browsing LLM agent as a proposer in the same format as our 17 human annotators and study whether it can produce challenging and well-defined problems.
When naively instructed, the machine-generated problems were either solvable by frontier models or ill-defined.
However, when we (i) provide hard human-written problems as few-shot exemplars or (ii) instruct the agent to target categorized failure modes identified in our analysis, the generated problems become substantially higher in quality and difficulty.
Specifically, the models achieve 0.00--26.00\% accuracy on the 100 machine-generated \textsc{Synthetic} problems, compared with 0.00--45.67\% on the corresponding subset of \textsc{K-BrowseComp-Verified}.
These results indicate that our generation method produces a challenging diagnostic split.


\section{Related Work}
\paragraph{Web browsing agents.}
Prior work established the foundation of evidence retrieval and reasoning, the core components of open-domain QA~\citep{kwiatkowski2019natural,yang2018hotpotqa,trivedi2022musique,press2023measuring}.  However, their evidence sources are typically predefined (\textit{e.g.}, Wikipedia), or LLMs operate over a fixed number of short turns, which became insufficient to evaluate frontier LLMs.
Hence, recent web browsing benchmarks test LLMs as agents that use tools (\textit{e.g.}, web search) over a longer horizon across multiple websites. 
For instance, BrowseComp~\citep{wei2025browsecomp} proposed a set of challenging questions that even humans with access to a web browser can't solve within 2 hours, and BrowseComp-ZH~\citep{zhou2025browsecompzh} extends the paradigm to utilizing multiple Chinese websites. 
In our work, we propose \textsc{K-BrowseComp}, which necessitates accessing multiple Korean websites. 
Beyond simply curating the question in Korean language, we consider a broad range of components that reflect the needs of Korean users (\textit{e.g.}, search conventions, local entities in Korean websites, and culturally grounded clues) that aren't covered by existing benchmarks.

\begin{figure*}[t!]
\centering
\includegraphics[width=0.99\linewidth]{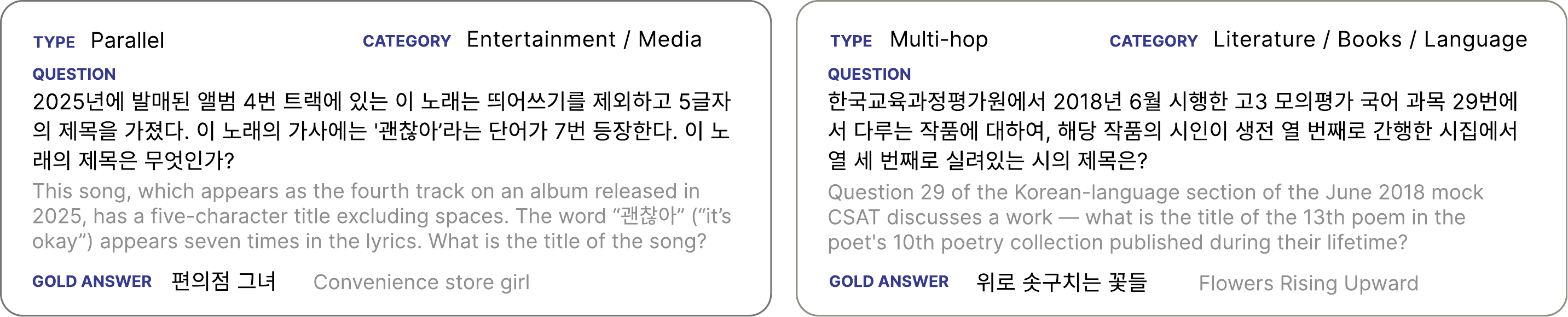}
\vspace{-2mm}
    \caption{\textbf{Examples of \textsc{K-BrowseComp} items.} The left example requires parallel-branching (\textit{i.e.}, gathering information from multiple websites) while the right example requires multi-hop reasoning (\textit{i.e.}, sequentially traversing through websites).}
    \label{fig:question_example}
\vspace{-2mm}
\end{figure*}

\paragraph{Korean and regional language evaluation.}
Korean LLM evaluation has mainly focused on static language understanding, factual knowledge, and reasoning. 
KorQuAD and KoBEST measure reading comprehension and core NLU skills, while KMMLU scales multitask knowledge evaluation for Korean~\citep{lim2019korquad, jang2022kobest,son2025kmmlu}. 
CLIcK~\citep{kim2024click} evaluates cultural and linguistic intelligence in Korean, and recent regional or multilingual benchmarks such as INCLUDE~\citep{romanou2025include} and MENLO~\citep{whitehouse2026menlo} emphasize locally grounded knowledge and native-like norms.
These resources are important for Korean language evaluation, but they usually do not require agents to search the web, maintain evidence state, or synthesize information across pages. 
\textsc{K-BrowseComp} fills this gap by converting Korean cultural, institutional, educational, geographic, and media knowledge into agentic browsing tasks.
This design evaluates whether a model can act as a useful Korean web agent, beyond answering questions from static benchmark distributions.

\paragraph{Synthetic task generation.}
A growing body of work uses LLMs to scale instruction data and evaluation data, including self-instruction, synthetic task generation, and adversarial example construction \citep{wang2023selfinstruct,xu2023wizardlm,nie2020adversarialnli}. At the same time, benchmark quality remains a central concern because model-generated tasks can be underspecified, too easy, or contaminated by highly indexed sources. Recent work on data contamination and benchmark generalization shows that evaluation results can overstate model ability when tasks are memorized or only require shallow generalization \citep{dong2024generalization}. 
\textsc{K-BrowseComp} contributes a complementary construction strategy for web-browsing tasks. It uses human-verified Korean browsing questions to identify recurring failure modes, then asks a browsing agent to generate new questions that target those modes. This pipeline uses the verifiability structure of browsing questions to scale Korean agentic evaluation while keeping the answer unique and publicly supported.

\begin{table*}[h]
\centering
\scriptsize
\setlength{\tabcolsep}{4pt}
\renewcommand{\arraystretch}{1.15}
\begin{tabular}{p{0.04\linewidth} p{0.28\linewidth} p{0.56\linewidth}}
\toprule
\textbf{Mode} & \textbf{Name} & \textbf{Definition} \\
\midrule
\textbf{F0} & Incomplete trajectory or malformed output 
& The model produces an incomplete trajectory, malformed output, or no valid final answer. \\

\textbf{F1} & Ineffective initial search direction 
& The model fails to choose a useful initial search strategy. \\

\textbf{F2} & Search-access structure failure 
& The model fails to access evidence hidden behind difficult page structures. \\

\textbf{F3} & Cross-source hopping failure 
& The model fails to connect evidence across weakly linked sources or entity contexts. \\

\textbf{F4} & Semi-structured parsing failure 
& The model misreads tables, lists, rankings, databases, or institutional pages. \\
\textbf{F5} & Search-result selection failure 
& The model retrieves relevant evidence but selects the wrong source or candidate. \\
\textbf{F6} & Sparse entity normalization failure 
& The model fails to resolve rare names, aliases, spelling variants, or historical names. \\

\textbf{F7} & Constraint-tracking failure 
& The model finds partial candidates but fails to satisfy all constraints. \\

\textbf{F8} & Intermediate reasoning failure 
& The model fails at date arithmetic, ordering, counting, comparison, or filtering. \\

\bottomrule
\end{tabular}
\caption{
\textbf{Failure-mode taxonomy.} We manually identify and label reasons why models fail on \textsc{K-BrowseComp-Verified}. These are later used as the source to construct the \textsc{Synthetic} subset of \textsc{K-BrowseComp}.}
\label{tab:failure_modes}
\vspace{-2mm}
\end{table*}

\section{\textsc{K-BrowseComp}}
\textsc{K-BrowseComp} is a Korean browsing benchmark designed to evaluate whether web browsing agents can retrieve hard-to-find public information grounded in Korean contexts to solve user queries.
As shown in Figure~\ref{fig:question_example}, each question is designed to have a single short answer that is stable over time and supported by publicly accessible web evidence.
The same figure illustrates the benchmark's two reasoning formats. \emph{Parallel-branching} questions require intersecting multiple independent constraints to identify a unique answer, while \emph{multi-hop} questions require using an intermediate finding to retrieve subsequent evidence. In the following sections, we describe the construction and validation of the \textsc{K-BrowseComp-Verified} subset (\S\ref{sec:K-BrowseComp-verified}) and the 100 \textsc{synthetic} split (\S\ref{sec:K-BrowseComp-llm}).

\subsection{\textsc{K-BrowseComp-Verified}}
\label{sec:K-BrowseComp-verified}

\paragraph{Annotator guidelines.}

We ask annotators to follow three rules when constructing \textsc{K-BrowseComp-Verified}.
First, each question should be grounded in Korean contexts and supported by public textual web evidence.
Second, each question should be difficult to answer through direct search, but easy to verify once the answer was found.
Third, each question should require either multi-hop reasoning or parallel constraint satisfaction, with at least four steps or constraints.
We prohibit annotators from using LLMs to create questions, and submitting questions that relies on private, paid, downloaded, or non-textual sources. Also, we require each item to have a unique answer (\textit{i.e.}, no questions that have multiple plausible answers) and temporally stable answer (\textit{i.e.}, answers that don't change over time). 
Further details are explained in Appendix~\ref{app:question_guidelines}.

\paragraph{Validation procedure.}

All finalized items are manually verified by the authors.
For each item, we check whether the gold answer, intermediate entities, and cited sources are recoverable from public web evidence.
Items with inaccessible, insufficient, or inconsistent evidence are returned to the original annotators for revision.
We also review each item for natural wording, temporal stability, and answer uniqueness.
When our baselines produce a concrete answer that differs from the gold answer, the case is manually inspected to determine whether the model finds a plausible alternative answer.
An item is retained only when the problem statement, gold answer, expected trajectory, source URLs, and checklist values are mutually consistent, and when the answer is unique.

\begin{figure}[t!]
\centering
\includegraphics[width=1.0\linewidth]{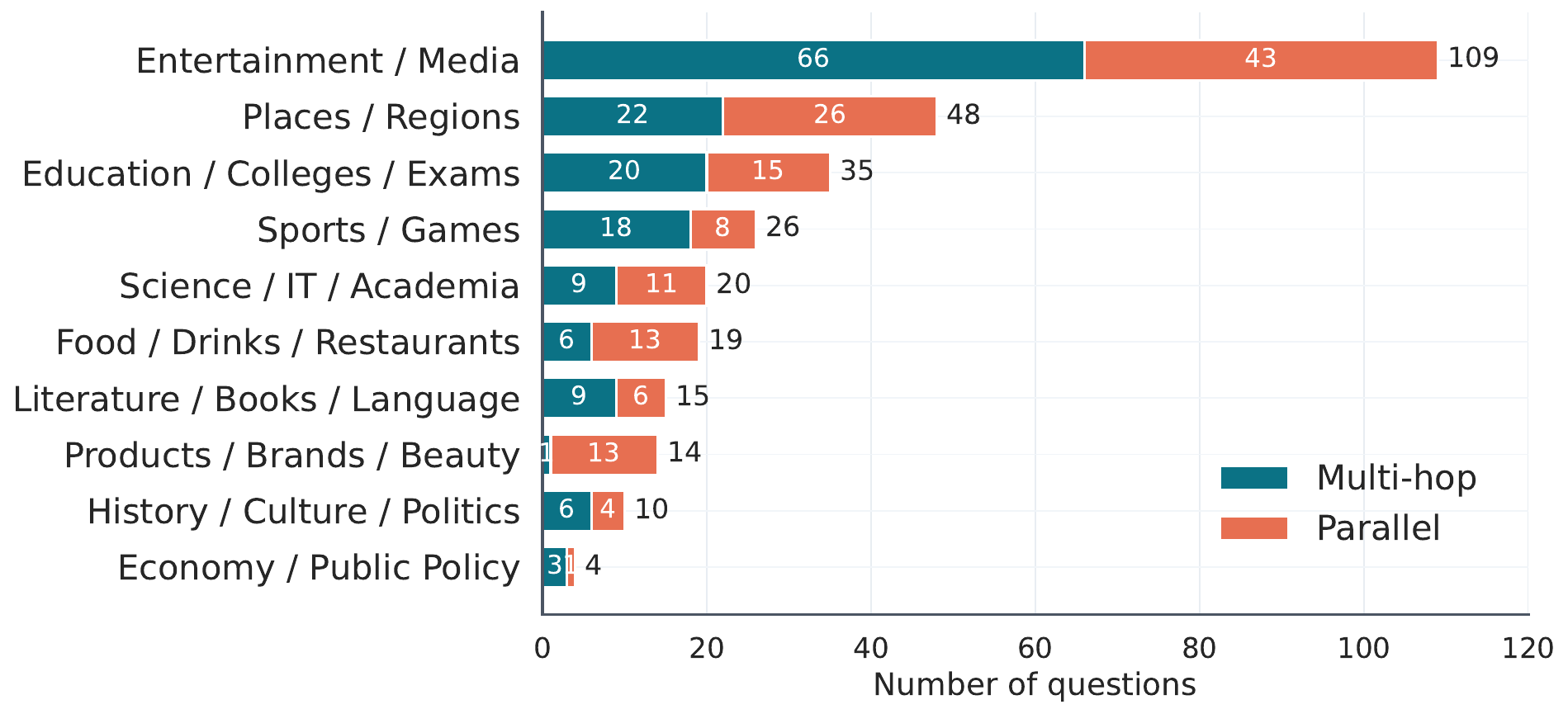}
\caption{
\textbf{Category distribution of \textsc{K-BrowseComp-Verified}.}
Bars show the number of questions in each category, decomposed by question type.
Numbers inside bars indicate the counts of \textit{multi-hop} and \textit{parallel-branching} questions, and numbers at the end of bars indicate category totals.
}
    \label{fig:dataset_distribution}
\vspace{-2mm}
\end{figure}

\paragraph{Dataset statistics.}

The verified subset consists of \textbf{300} questions, where 160 are \textit{multi-hop} questions (53.3\%) and 140 are \textit{parallel-branching} questions (46.7\%).
Figure~\ref{fig:dataset_distribution} illustrates the category and reasoning-format distribution.
The largest category is \textit{Entertainment and Media}, with 109 questions (36.33\%).
The second largest category is \textit{Transportation, Places, and Regions}, with 48 questions (16.00\%).
The remaining questions cover a range of categories, including \textit{education, sports, science, literature, products, history}, and \textit{public policy}.

\paragraph{Trajectory-level failure taxonomy.}\label{sec:failure_taxonomy}
Each \textsc{K-BrowseComp} item requires an extended information-seeking process: forming useful search queries, retrieving candidate evidence, selecting relevant results, interpreting semi-structured source metadata, combining information across multiple steps or parallel branches, and deriving the final answer.
To identify where errors occur, we define a trajectory-level failure-mode taxonomy, summarized in Table~\ref{tab:failure_modes}. 
The taxonomy is constructed through manual inspection of recurring model errors.
In Appendix~\ref{app:failure_modes_details}, we provide examples of each failure mode.

\subsection{Generating \textsc{synthetic} web-browsing problems with browsing agents}
\label{sec:K-BrowseComp-llm}

\paragraph{Motivation} While constructing K-BrowseComp-Verified, we identified which Korean websites web browsing agents fail to retrieve information from, as well as the behaviors they exhibit when they fail. However, asking humans to construct such problems is costly and time-consuming. Inspired by recent approaches to generating synthetic problems~\citep{kim2025biggen,liu2025spice,kulikov2026autodata}, we therefore ask whether we can leverage this failure taxonomy, together with the identified websites, to have an AI agent create equally challenging web browsing problems.

\paragraph{Generation pipeline.}
To generate \textsc{synthetic} problems, \textsc{K-BrowseComp-Verified} provides both seed pages and a failure taxonomy. Its gold source pages are used to collect related Korean web pages, while observed baseline failures define the nine \emph{failure modes} in Table~\ref{tab:failure_modes}. For generation, we target F1--F8 and exclude F0, since malformed or incomplete trajectories are not a reusable content-level failure for question construction.
Each seed page is provided to Claude Code (\texttt{claude-opus-4.7} at maximum effort) and is tasked with drafting a single adversarial question. Specifically, it opens the web page and constructs a Korean question \emph{backwards} (i.e., as opposed to opening a web page after reading a question) so that reaching it requires multi-hop or parallel-constraint retrieval, as in the verified set. The question withholds the answer and any paraphrase of it, the source URL and domain, and the page's most identifying entity name, and targets a designated failure mode. The agent refines the question over four iterations (draft $\rightarrow$ test $\rightarrow$ revise) until it satisfies the filtering rules.

\paragraph{Filtering rules.}
Each candidate question passes three sequential filters, and failing any one returns the agent to revision. The first tests \emph{searchability}: the agent issues several web queries and checks whether the gold answer already appears among the results; if so, the question is too easy and is rewritten to target a less exposed detail. The second tests \emph{well-formedness}: a reference solver receives the full source page together with the question and must recover the answer, confirming that it is uniquely and faithfully extractable from the page rather than ambiguous or absent. The third tests \emph{adversarial difficulty}: a search-only solver, granted web search but no direct access to the page, attempts the question, and the candidate is retained only if every target model (\texttt{gpt-5.4-mini} and \texttt{gemini-3-flash-preview}) fails, where failure includes both an incorrect answer and an explicit abstention. Acceptance further requires that each model's failure be attributable to one of the eight failure modes, F1--F8; the agent labels the realized mode honestly, re-assigning it from the intended one when the observed failure differs, for instance recording an unreachable-page failure as F2 rather than disguising it as a parsing failure.

\paragraph{Dataset statistics.}
Among the 268 candidate questions we generated, 100 problems were not filtered out (37.3\% yield). Among the accepted questions, 55, 32, 10, and 3 are obtained on the first through fourth iteration, respectively. Each accepted question is labeled with the failure mode(s) it elicits, distributed as F4 (59), F7 (21), F2 (14), F8 (13), F3 (13), F6 (8), F5 (6), and F1 (1), with semi-structured parsing and constraint accumulation being the most frequently exploited weaknesses. Rejections (\textit{i.e.}, problems that were filtered out) are dominated by seed pages that are already fully indexed: for the majority of the 168 rejected items the gold answer surfaces directly in search (\textit{i.e.}, \textit{searchability}). Also, 66 are solvable by either \texttt{gemini-3-flash-preview} or \texttt{gpt-5.4-mini}. In the subsequent sections, we use the 100 problems that were not filtered.

\paragraph{\textsc{Synthetic} split diagnostics.}
We analyze the generated problems as a complementary stress split.
The \textsc{synthetic} split preserves the reasoning-format balance of \textsc{K-BrowseComp-Verified}, with 53.0\% multi-hop and 47.0\% parallel examples.
Its main differences appear in category composition and question length.
The share of \textit{Entertainment and Media} drops from 36.3\% to 9.0\%, while the share of \textit{Science, IT, and Academia} rises from 6.7\% to 33.0\%.
The questions are also longer, with mean length increasing from 174.46 to 248.40 characters.
We further embed all questions with a multilingual sentence-transformer model and test whether the two splits can be separated from question text alone.
A simple classifier separates the two splits well, with an ROC AUC of 0.8873.
These results suggest that the \textsc{synthetic} split preserves the reasoning format, but differs in domain and surface profile.
We therefore report it separately from \textsc{K-BrowseComp-Verified} in Section~\ref{sec:result_synthetic} and provide further analyses in Appendix~\ref{app:synthetic_diagnostics}.

\section{Experimental Setup}\label{sec:experimental_setting}

\paragraph{Baselines.}
We evaluate proprietary models and open-weight models on \textsc{K-BrowseComp-Verified}.
The proprietary baselines include \texttt{GPT-5.5}, \texttt{GPT-5.4-mini}, and \texttt{Gemini-3.1-Flash-Lite}~\cite{openai2026gpt55,google2026gemini31flashlite}.
The open-weight baselines include \texttt{DeepSeek-V4-Pro}, \texttt{GLM-5.1}, \texttt{Qwen3.6-35B-A3B}, and \texttt{Gemma-4-31B-it}~\cite{deepseekai2026deepseekv4,zai2026glm51,qwen2026qwen36,google2026gemma4}.
We also evaluate Korean open-weight models, including \texttt{K-EXAONE-236B-A23B}, \texttt{A.X-4.0}, \texttt{HyperCLOVAX-SEED-Think-32B}, and \texttt{Kanana-2-30B-A3B-Thinking-2601} \cite{lgai2026kexaone,skt2025ax40,team2026hyperclova,kakao2026kanana2}.

\paragraph{Evaluation protocol.}
We build our browsing-agent evaluation on the \texttt{search\_evals} framework~\citep{2025pplxsearchevals}.
We use the \textit{deep-research agent} and set \textit{Perplexity Search} as the search backend for evaluation.
For comparability across models, each agent is given a budget of 10 search calls per question, which is the default setting of the \mbox{\texttt{search\_evals}} framework.
Each model is evaluated once on each of the 300 questions in the verified subset.
We use \texttt{GPT-5.4-mini} to extract the final answer from each model response and match it against the gold answer, following BrowseComp~\cite{wei2025browsecomp}.
We report the single-run accuracy, which corresponds to pass@1 in this setting.


\begin{table}[t]
\centering
\scriptsize
\setlength{\tabcolsep}{3.2pt}
\resizebox{\columnwidth}{!}{%
\begin{tabular}{lcccc}
\toprule
 & & \multicolumn{2}{c}{\textbf{Verified subset}} & \textbf{Synthetic subset} \\
\cmidrule(lr){3-4} \cmidrule(lr){5-5}
\textbf{Model} & \textbf{Access} & \textbf{Pass@1 Acc.} & \textbf{Calib. Err.} & \textbf{Pass@1 Acc.} \\
\midrule
GPT-5.5 & Closed & 45.67 & 31.86 & 26.00 \\
GPT-5.4-mini & Closed & 30.67 & 37.88 & N/A$^{\dagger}$ \\
DeepSeek-V4-Pro & Open & 30.00 & 17.72 & 22.00 \\
GLM-5.1 & Open & 30.67 & 27.07 & 19.00 \\
Qwen3.6-35B-A3B & Open & 12.00 & 47.89 & 15.00 \\
Gemini-3.1-Flash-Lite & Closed & 11.33 & 56.55 & 11.00 \\
Gemma-4-31B-it & Open & 23.33 & 23.66 & 17.00 \\
\midrule
K-EXAONE-236B-A23B & Open & 10.33 & 24.09 & 13.00 \\
A.X-4.0 & Open & 5.33 & 47.89 & 1.00 \\
HCX-SEED-Think-32B & Open & 2.33 & 77.37 & 2.00 \\
Kanana-2-30B-A3B-Think & Open & -- & -- & -- \\
\bottomrule
\end{tabular}
}
\caption{
\textbf{Performance on \textsc{K-BrowseComp-Verified} and the \textsc{synthetic} split.}
Pass@1 accuracy (\%) and calibration error (\%) are computed on the 300-question \textsc{K-BrowseComp-Verified} subset.
\textsc{Synthetic} accuracy is computed on the 100-question diagnostic synthetic split for the completed subset of models, and is not pooled with the verified score.
All runs use the same external retrieval pipeline with the Perplexity Search API.
$\dagger$ Note that GPT-5.4-mini scores N/A since we adversarially created problems.
}
\label{tab:main_results}
\end{table}

\section{Experimental Results}\label{experimental_results}

\subsection{Main results}
Table~\ref{tab:main_results} reports the performance of closed and open models on \textsc{K-BrowseComp-Verified}.
\texttt{GPT-5.5} achieves the highest accuracy, with 45.67\%.
Among the remaining models, \texttt{GPT-5.4-mini} and \texttt{GLM-5.1} both reach 30.67\%, while \texttt{DeepSeek-V4-Pro} obtains 30.00\%.
These results show that the benchmark remains challenging even for recent high-performing models.

Smaller open-weight models show a wide performance range.
\texttt{Gemma-4-31B-IT} reaches 23.33\%, outperforming \texttt{Qwen3.6-35B-A3B} at 12.00\%.
Korean open models score lower in this setting, with \texttt{K-EXAONE-236B-A23B} at 10.33\%, \texttt{A.X-4.0} at 5.33\%, and \texttt{HyperCLOVAX-SEED-Think-32B} at 2.33\%. 
The ``--'' entries indicate that \texttt{Kanana-2-30B-A3B-Thinking-2601} frequently produced degenerate character sequences or malformed tool calls, preventing valid browsing trajectories under our harness.

Following BrowseComp and BrowseComp-ZH~\citep{wei2025browsecomp,zhou2025browsecompzh}, we report expected calibration error over five equal-width confidence bins, computed as the weighted average gap between mean confidence and empirical accuracy. Lower values indicate better alignment between confidence and correctness.

\subsection{Results on Synthetic Splits}\label{sec:result_synthetic}
The last column of Table~\ref{tab:main_results} reports performance on the 100-question \textsc{Synthetic} split. We evaluate this split using the same browsing-agent harness and grading protocol as in Section~\ref{sec:experimental_setting}.
Accuracy remains low across all evaluated models, ranging from 1.00\% to 26.00\%, and no model exceeds 30.00\% on this split. Even the strongest models on the verified subset remain substantially below their verified-subset accuracy, with \texttt{GPT-5.5}, \texttt{DeepSeek-V4-Pro}, and \texttt{GLM-5.1} scoring 26.00\%, 22.00\%, and 19.00\%, respectively. These results suggest that the \textsc{Synthetic} split exposes persistent browsing failures under the same evaluation protocol.
Because failure-based filtering is part of the construction process, we use the \textsc{synthetic} split as a diagnostic stress test and do not pool it with the verified score. We treat the \texttt{GPT-5.4-mini} result separately here because this model was used during adversarial filtering.

\begin{figure}[!t]
    \centering
    \includegraphics[width=0.99\linewidth]{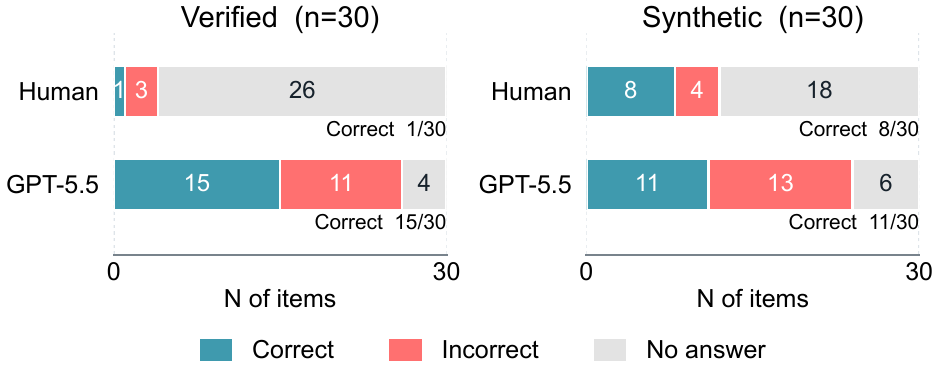}
    \caption{\textbf{Outcome distributions for the time-bounded human baseline and item-matched \texttt{GPT-5.5}.} Each panel contains 30 items from the indicated split.}
    \label{fig:human-baseline}
\end{figure}


\begin{figure*}[t!]
\centering
\includegraphics[width=0.95\linewidth]{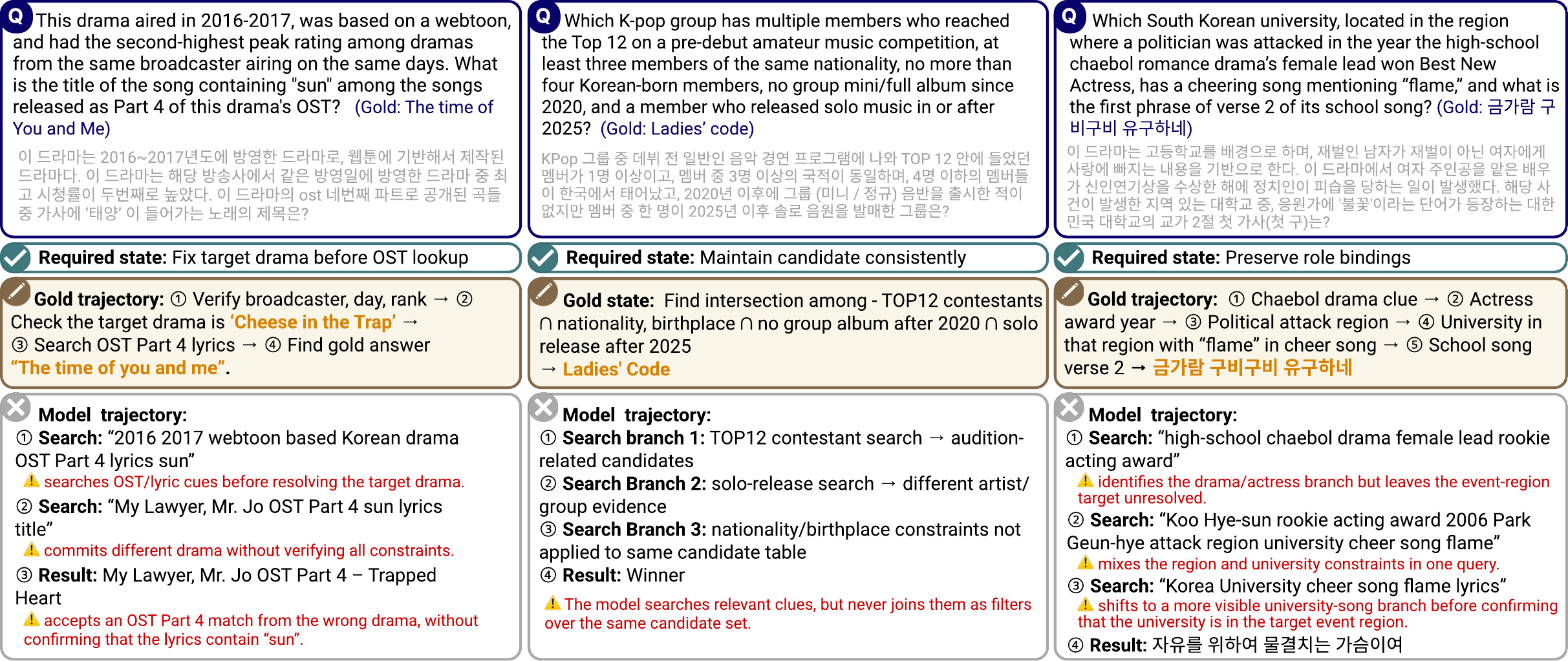}
\vspace{-2mm}
\caption{
\textbf{Representative trajectory-level failures in \textsc{K-BrowseComp}.}
Each panel contrasts the required intermediate state with the model trajectory.
The examples illustrate three recurring post-retrieval failures: \emph{candidate capture}, \emph{unmerged evidence branches}, and \emph{misbound evidence chains}.
}
\label{fig:trajectory_cases}
\vspace{-3mm}
\end{figure*}

\subsection{Time-bounded human baseline}
\label{sec:human_baseline}

We randomly sample 60 items, 30 from each split, and assign each to one of four native Korean speakers for one 30-minute attempt. Participants may use search engines but not LLM tools. Failure to answer within 30 minutes is labeled no answer. For \texttt{GPT-5.5}, explicit abstentions such as \textit{I don't know} are labeled no answer, while other non-gold responses are incorrect. Both outcomes count as non-correct for accuracy.

For an item-matched comparison, we use the existing \texttt{GPT-5.5} outputs. Figure~\ref{fig:human-baseline} shows the outcome distributions. Humans solve 9/60 items (15.00\%), compared with 26/60 (43.33\%) for \texttt{GPT-5.5}. Humans solve 1 and 8 items on the \textsc{Verified} and \textsc{Synthetic} samples, while \texttt{GPT-5.5} solves 15 and 11. Human outcomes include 7 incorrect and 44 no-answer cases, while the model produces 24 and 10, respectively. The 44 human no-answer cases exhaust the time limit and total 22 person-hours. Higher human accuracy on \textsc{Synthetic} split is consistent with its construction as a model-oriented stress test. Appendix~\ref{app:time_bounded_human_baseline} provides the participant, assignment, and scoring details.

\section{Analysis}

\subsection{Trajectory-Level Failure Patterns}\label{sec:trajectory_failures}
Whereas Table~\ref{tab:failure_modes} catalogs failure modes at the level of individual reasoning or search steps, here we examine how a subset of these modes manifest as recurring \textit{trajectory-level dynamics}: patterns in which step-level errors compose across many steps after partially relevant evidence has already been retrieved. Many errors arise after models have already found partially relevant evidence, and these errors often come from losing track of candidates, constraints, and evidence across multiple steps. An illustration is shown in Figure~\ref{fig:trajectory_cases}.

\paragraph{F5+F7: Search-result selection \& constraint-tracking failure (candidate capture).}
This failure occurs when the model commits to a plausible entity before all upstream constraints have been verified.
After this early commitment, later searches become confirmatory: the model searches within the local evidence space of the selected candidate and treats unresolved constraints as satisfied. The final answer can appear locally supported, while still violating the full question. In the left panel of Figure~\ref{fig:trajectory_cases}, the model should first identify the target drama before searching its OST lyrics, but instead starts from OST and lyric cues and anchors on \textit{My Lawyer, Mr. Jo}. It then returns \textit{Trapped Heart}, an OST Part~4 song from the wrong drama, illustrating how premature candidate commitment can mislead. See Appendix~\ref{app:candidate_capture} for additional examples.

\paragraph{F7: Constraint-tracking failure (unmerged evidence branches).}
In this pattern, the model searches for several relevant constraints, but each query creates a \textit{separate evidence branch}.
The branches are never converted into filters over a shared candidate set. The trajectory may appear systematic because it touches many clues, but the candidate table remains inconsistent across steps.

The middle panel of Figure~\ref{fig:trajectory_cases} shows this pattern.
The question asks for a K-pop group satisfying parallel constraints on pre-debut competition history, member nationality, and album release status.
The gold state is the intersection of these constraints, which yields \textit{Ladies' Code}.
However, the model searches these clues separately, applies them to different artists or groups, and returns \textit{Winner}, even though it violates the album-release constraint.
This error shows that the retrieved clues were not used to derive the answer consistently.

\paragraph{F3: Cross-source hopping failure (misbound evidence chains).}
In this pattern, the model follows a plausible search sequence, but binds an intermediate result to the wrong role. This is most visible when the question changes entity type across steps.
If the region, institution, song, and lyric roles are not preserved, a high-visibility entity can replace the intended target.
The right panel of Figure~\ref{fig:trajectory_cases} illustrates this failure. The gold path must preserve a chain from the drama to the actress's award year, the political attack region, a university in that region, and the first phrase of verse~2 of its school song. The model identifies part of the drama and actress branch, but leaves the event region unresolved. It then folds the award, event, university, and cheer-song clues into a broad query, shifts to a visible university-song branch, and returns a plausible phrase from the wrong institution.

\paragraph{F0: Incomplete trajectory or malformed output (answer finalization).}
Certain models begin a reasonable browsing trajectory but do not stabilize it into a well-formed final answer.
This includes incomplete trajectories, unstable intermediate conclusions, excessive uncertainty, or final responses that do not follow the expected answer format.
These cases show that browsing performance depends not only on retrieval and reasoning, but also on controlled finalization.
The model must know when the evidence is sufficient, when more search is needed, and how to produce a concise exact answer.

\paragraph{Takeaway.}
Together, these patterns show that many browsing failures emerge after required evidence has been retrieved, and that F0, F3, F5, and F7 are the dominant failure modes once retrieval has succeeded.
This suggests that progress on \textsc{K-BrowseComp} depends on stronger mechanisms for maintaining candidates, constraints, role bindings, and final-answer state across multiple turns.

\begin{table}[t]
\centering
\scriptsize
\setlength{\tabcolsep}{2.5pt}
\begin{tabular}{@{}lccc@{}}
\toprule
\textbf{Model} &
\shortstack{\textbf{Correct $n$ /}\\\textbf{avg. calls}} &
\shortstack{\textbf{Wrong $n$ /}\\\textbf{avg. calls}} &
\shortstack{\textbf{$\Delta$}\\\textbf{calls}} \\
\midrule
GPT-5.5        & 137 / 7.09 & 163 / 9.21 & +2.13 \\
GPT-5.4-mini   & 92 / 7.88  & 208 / 9.51 & +1.63 \\
Gemini-3.1-Flash-Lite  & 34 / 5.26  & 266 / 6.73 & +1.47 \\
DeepSeek-V4-Pro & 90 / 7.47  & 210 / 9.80 & +2.33 \\
GLM-5.1        & 92 / 8.23  & 208 / 9.38 & +1.15 \\
Gemma-4-31B-it    & 70 / 5.20  & 230 / 8.10 & +2.90 \\
Qwen3.6-35B-A3B    & 36 / 8.11  & 264 / 9.69 & +1.58 \\
K-EXAONE-236B-A23B       & 31 / 8.35  & 269 / 9.45 & +1.10 \\
A.X-4.0        & 16 / 2.38  & 284 / 1.43 & -0.95 \\
HyperCLOVAX-SEED-Think-32B   & 7 / 6.71   & 293 / 6.84 & +0.13 \\
\bottomrule
\end{tabular}
\caption{
\textbf{Search-call usage by outcome.} Each cell reports the number of trials and the average number of search calls on \textsc{K-BrowseComp-Verified}. $\Delta$ is the wrong-trial average minus the correct-trial average. All runs use a 10-call budget.
}
\label{tab:search_usage}
\vspace{-1mm}
\end{table}

\begin{table}[t]
\centering
\scriptsize
\setlength{\tabcolsep}{3.2pt}
\begin{tabular}{@{}lrrr@{}}
\toprule
\textbf{Model}
& \textbf{Perplexity-10}
& \textbf{Perplexity-20}
& \textbf{Exa-10} \\
\midrule
Gemma-4-31B-it
& 29.0 & 28.0 & 17.0 \\
K-EXAONE-236B-A23B
& 12.0 & 16.0 & 15.0 \\
\bottomrule
\end{tabular}
\caption{
\textbf{Retrieval-configuration sensitivity.}
Pass@1 accuracy (\%) on the fixed 100-item sample containing
80 \textsc{Verified} and 20 \textsc{Synthetic} items. The suffix denotes the search-call budget.
}
\label{tab:retrieval_sensitivity}
\vspace{-1mm}
\end{table}

\subsection{Search Effort and Failure Persistence}
\label{sec:search_sensitivity}

One possible explanation for the low scores is that the default 10-call budget restricts evidence acquisition. We first examine search-call usage under the primary \texttt{Perplexity} Search configuration. Table~\ref{tab:search_usage} compares usage between correct and incorrect trials. For most models, incorrect trials use more search calls than correct trials and often approach the available budget.

The difference is particularly large for \texttt{GPT-5.5} (7.09 vs.\ 9.21), \texttt{DeepSeek-V4-Pro} (7.47 vs.\ 9.80), \texttt{Gemma-4-31B-it} (5.20 vs.\ 8.10), and \texttt{Qwen3.6-35B-A3B} (8.11 vs.\ 9.69). In contrast, \texttt{A.X-4.0} uses fewer calls on incorrect trials, which is consistent with shallow search or premature stopping. \texttt{HyperCLOVAX-SEED-Think-32B} shows little separation between correct and incorrect trials, consistent with failures in trajectory completion or final-answer stabilization.

These patterns indicate that failures often persist despite additional search within the tested budget. In conjunction with the trajectory analysis above, they suggest that many models struggle to maintain candidates, constraints, and entity roles across the searches they perform. However, the association between call usage and outcome does not show that the available budget is sufficient or that increasing it would not improve performance.

\paragraph{Retrieval-configuration sensitivity.} We therefore conduct a controlled sensitivity experiment using a fixed sample of 100 questions, consisting of 80 \textsc{Verified} and 20 \textsc{Synthetic} items. We evaluate \texttt{Gemma-4-31B-it} and \texttt{K-EXAONE-236B-A23B} under three retrieval configurations. The base configuration uses \textit{Perplexity} search backend with a 10-call budget. We compare it with \textit{Perplexity} using 20 calls and \textit{Exa} using 10 calls. 
Within each model comparison, we hold the agent logic, prompt, answer extraction, and grading procedure fixed.

Increasing the \textit{Perplexity} budget from 10 to 20 calls changes accuracy by $-1.0\%$ for \texttt{Gemma-4-31B-it} and $+4.0\%$ for \texttt{K-EXAONE-236B-A23B}. Replacing \textit{Perplexity} with \textit{Exa} changes accuracy by $-12.0\%$ and $+3.0\%$, respectively. The effects differ across the two models and do not show a uniform benefit from increasing the call budget or switching the retrieval backend. Overall, search volume or search backend are a weak predictor of success. A stronger signal is whether the model can maintain candidate, constraint, and role state across searches it performs.




\subsection{Failures in Korean Open-Weight Models}

A benchmark should not merely report low scores but inform development. The Korean open-weight models we evaluate underperform global counterparts by a wide margin (Table~\ref{tab:main_results}); we provide a model-level diagnosis to guide future iterations.

\paragraph{\texttt{A.X-4.0}.} Built on Qwen2.5~\citep{qwen2025qwen25technicalreport}, the model often commits to plausible Korean web snippets before all constraints are verified, consistent with its unusually low search usage in Table~\ref{tab:search_usage}. Korean-targeted post-training has not, under this evaluation setting, translated into trajectory-level state maintenance (Appendix~\ref{app:korean_model_trajectory_analysis}).

\paragraph{\texttt{K-EXAONE-236B-A23B}.} Despite its scale, the model frequently loses the intermediate entity chain after relevant retrieval, with later queries drifting from the target entity (Appendix~\ref{app:korean_model_trajectory_analysis}). Parameter count alone does not appear to substitute for long-horizon state tracking.

\paragraph{\texttt{HyperCLOVAX-SEED-Think-32B}.} The model initiates reasonable browsing trajectories but rarely converges to a well-formed final answer. Its search counts on correct and incorrect trials are nearly identical (Table~\ref{tab:search_usage}), suggesting the bottleneck lies in finalization, not retrieval effort.

\paragraph{\texttt{Kanana-2-30B-A3B-Thinking-2601}.} The model does not produce completed browsing runs under the \texttt{search$\_$eval} harness, often emitting tool-call objects that violate the protocol. We read this as incomplete adaptation to the tool-use paradigm instead of a content-level limitation.

\paragraph{Takeaway.} Across models, Korean open-weight LLMs show no clear advantage on \textsc{K-BrowseComp} despite its Korean-centric design, and the gap to closed frontier APIs remains substantial. The bottlenecks we observe are stage-specific (\textit{e.g.}, tool-call protocol, post-retrieval state, and answer finalization) rather than reducible to a single cause. We position \textsc{K-BrowseComp} as a shared diagnostic target to encourage prioritization of browsing-agent capabilities alongside static-benchmark performance.

\subsection{Trajectory Diagnostics on the \textsc{Synthetic} Split}

A trajectory review helps explain why the \textsc{Synthetic} split remains difficult.
As reported in Section~\ref{sec:K-BrowseComp-llm}, accepted \textsc{Synthetic} questions concentrate on modes that the pipeline can instantiate reliably, especially semi-structured parsing (F4) and constraint tracking (F7).
The completed runs show that these targeted operations also reappear as solving failures.

Models often reach the right source family or intermediate entity, but fail at the page-level or candidate-level state needed for the final answer.
In a repository metadata question (Appendix~\ref{app:synthetic_trajectory_diagnostics} Figure~\ref{fig:synthetic_filesize_failure}), the model reaches the KOPRI repository source family, but extracts a nearby incorrect file-size value instead of the target PDF metadata field.
In a KBO record question (Appendix~\ref{app:synthetic_trajectory_diagnostics} Figure~\ref{fig:synthetic_kbo_constraint_failure}), the model identifies 안우진 (\textit{An Woo-jin}) from the 2022 Kiwoom Heroes pitching clues, but does not maintain a stable table of his 2026 appearances and opponent AVG values before selecting the final date.
These cases show that the main bottleneck is not broad retrieval. The \textsc{Synthetic} split instead demonstrates whether models can preserve the correct source or candidate state long enough to extract an exact value or enforce the final comparison.

\section{Conclusion}
We introduce \textsc{K-BrowseComp}, a Korean web-browsing agent benchmark with a 300-question human-verified subset and a 100-question \textsc{Synthetic} split. Even strong frontier models achieve low scores, and Korean open-weight models lag substantially behind global counterparts. The \textsc{Synthetic} split is similarly difficult, indicating that failure-mode-targeted generation, paired with verification and filtering, produces useful diagnostic items.
Our analysis shows that many failures occur after models retrieve relevant Korean web evidence: they often fail to maintain candidates, constraints, source pointers, or final-answer state across the trajectory. Progress on Korean browsing therefore requires stronger trajectory-level state maintenance, not only broader language coverage or larger model scale. We release \textsc{K-BrowseComp} as a target for building reliable web-browsing agents that function for Korean web environments.

\section*{Acknowledgements}
This research was supported by the “Open Source AI·SW Developer and Community Support Program” funded by the Ministry of Science and ICT (MSIT), Republic of Korea.

\section*{Limitations}
While \textsc{K-BrowseComp} provides a first step toward evaluating Korean web-browsing agents, several limitations remain. First, the verified set is still modest in scale and uneven in domain coverage: it contains 300 human-written items, with a substantial share concentrated in entertainment/media and place-related queries. Thus, the benchmark may not fully represent the breadth of Korean web-search use cases. Second, model performance is measured under a single browsing harness, search backend, search-call budget, and pass@1 setting, which may conflate model capability with retrieval coverage, tool-interface reliability, and evaluation protocol choices. Third, although the \textsc{Synthetic} split is useful as a diagnostic stress test, it differs from the verified set in surface form and domain composition. Finally, because web evidence and search rankings change over time, continued revalidation will be necessary to preserve answer uniqueness, accessibility, and reproducibility.


\bibliography{anthology,custom}
\clearpage


\appendix

\begin{figure*}[t!]
\centering
\includegraphics[width=0.8\linewidth]   {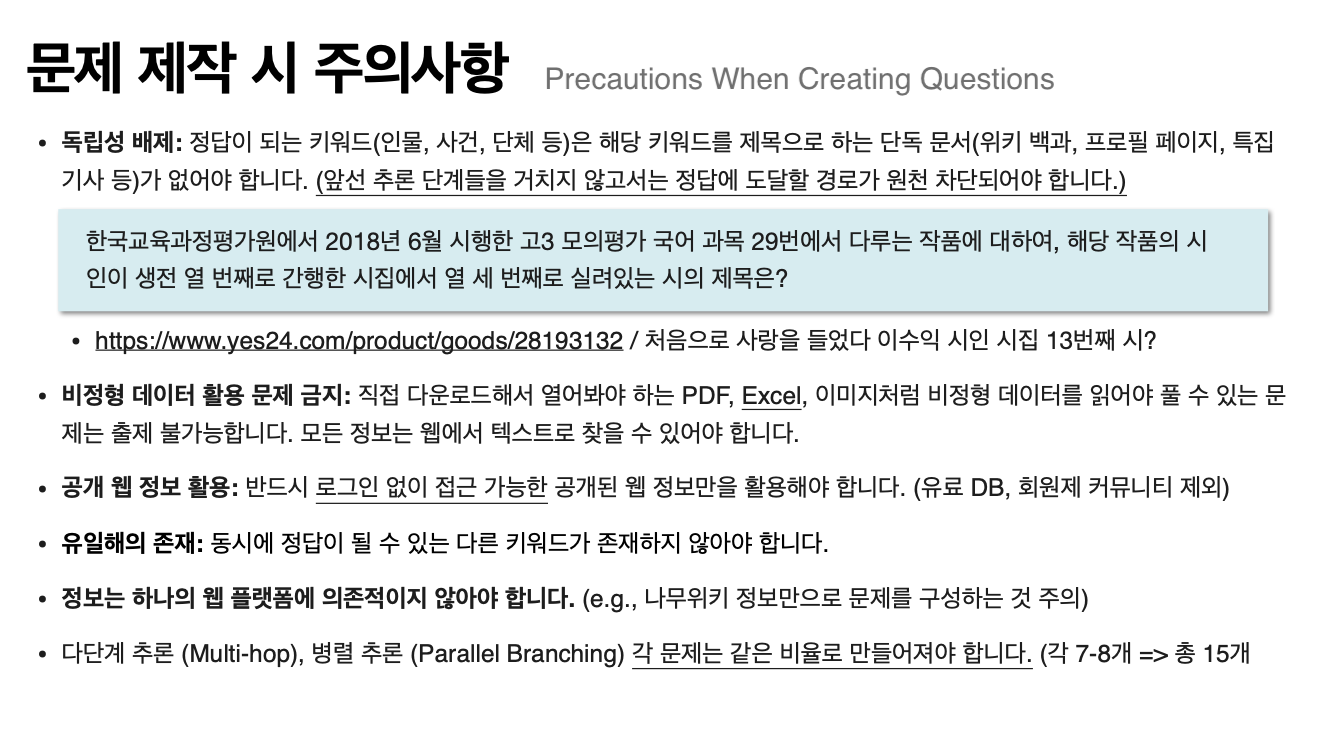}
\caption{
\textbf{Excerpt from the written instructions provided to contributors for constructing \textsc{K-BrowseComp-Verified} questions.}
The guide summarizes the main exclusion and validation rules: answer keywords should not be directly revealed by standalone documents, required evidence must come from publicly accessible textual web sources, non-textual artifacts such as PDFs, spreadsheets, and images are excluded, each question must have a unique answer, evidence should not depend on a single web platform, and multi-hop and parallel-branching items should be balanced.
}
    \label{fig:annotator_instruction}
\end{figure*}

\section{Dataset Construction} \label{app:question_guidelines}

\paragraph{Contributor instructions.}

Contributors were given a written guide that specified the goal, item format, and exclusion rules for \textsc{K-BrowseComp-Verified}.
Figure~\ref{fig:annotator_instruction} shows an excerpt of the instructions used during question construction.
They were asked to write difficult fact-seeking questions that require web browsing and evidence synthesis.
Each item had to be grounded in Korean contexts and had to have a short answer supported by public web evidence.
Contributors were encouraged to test draft questions with web-enabled models to check difficulty, but they were instructed not to use LLMs to generate or revise the questions.

\paragraph{Korean grounding.}

Each question had to contain Korean-specific information in the question, the answer, or the intended evidence path.
This included Korean institutions, local places, public transportation, educational materials, cultural artifacts, media, historical facts, or expressions commonly used in Korea.
The grounding requirement was intended to test browsing ability in Korean information environments, including sources whose structure, terminology, and search conventions differ from English web resources.

\paragraph{Reasoning formats.}

Each question followed one of two formats.
In \textit{multi-hop} questions, the solver first identifies an intermediate entity or value, then uses it to retrieve later evidence and reach the final answer.
In \textit{parallel-branching} questions, the solver evaluates several independent constraints and identifies the unique entity satisfying all of them.
Multi-hop items required at least four reasoning steps.
Parallel-branching items required at least four independent constraints.
Contributors were asked to balance the two formats across their submitted items.

\paragraph{Difficulty criteria.}

Questions were designed to be difficult to answer through direct keyword search.
Contributors typically began with a verified target fact, selected attributes with a large search space, and then wrote an inverted question that required recovering the target through search.
The answer and intermediate steps had to be easy to verify once the correct path was found.
At the same time, the answer keyword should not be the title of a standalone document that reveals the answer without following the intended reasoning path.

\paragraph{Evidence requirements.}

All required evidence had to be available from public web pages.
Sources requiring login, payment, private databases, or membership access were not allowed.
Items were also excluded when solving required downloading or interpreting PDFs, spreadsheets, images, or other non-textual artifacts.
Contributors were discouraged from relying on a single web platform as the sole basis for an item.
When possible, intermediate facts were supported by multiple independent sources.

\paragraph{Answer constraints.}

Each item had to have a unique final answer under the stated conditions.
Contributors were asked to add constraints when another answer could plausibly satisfy the question.
Items involving time-sensitive facts had to include a reference date in the question.
During review, questions were checked for ambiguous wording, unstable clues, and plausible alternative answers.

\paragraph{Submission format.}
Each item is submitted in a structured JSON format containing the problem statement, gold answer, expected solving trajectory, source URLs, checklist values for key intermediate slots, Korean-specific keywords, and a short rationale.
The checklist is used to check whether the evidence path is valid in our validation procedure.
The rationale explained why the item required Korean-context browsing and why direct search was unlikely to be sufficient.

\paragraph{Revision and acceptance.}

Submitted items were manually reviewed for evidence accessibility, source consistency, wording clarity, answer uniqueness, and temporal stability.
During this review, we also checked submissions for private personal information, sensitive personal data, and offensive content, and revised or removed items that failed this screen.
Items were returned to the original contributor when the evidence was inaccessible, insufficient, inconsistent, or dependent on excluded source types.
Items were also revised or removed when a baseline web-enabled model produced a different concrete answer that was judged to be a plausible alternative.
An item was accepted only after the problem statement, gold answer, evidence path, source URLs, and checklist values were mutually consistent.

\paragraph{Recruitment and Payment}
Non-researcher participants of this study were recruited via a social media post and the remaining participants were authors of this paper. Non-researcher participants were paid ₩100,000 (approximately USD 60–70) per 10–15 instances crafted, which corresponds to roughly 4 hours of work and exceeds South Korea's 2025 minimum hourly wage of ₩10,030. Author-participants did not receive separate compensation.

\paragraph{Data Consent}
All participants were aware that the annotations would be part of constructing the \textsc{K-BrowseComp} benchmark.

\paragraph{Artifact release and licensing.}
We will release \textsc{K-BrowseComp}, including the verified and \textsc{synthetic} questions, source URLs, expected trajectories, checklist values, and evaluation code, to support research and evaluation.
The \textsc{K-BrowseComp} dataset and evaluation code will both be released under the MIT License.
The license applies only to benchmark items, metadata, and code created by the authors and contributors; it does not grant rights to third-party web pages linked as evidence, whose contents remain governed by their original providers' terms.
We provide source URLs for verification but do not redistribute the full contents of linked web pages.


\begin{figure*}[h]
\centering
\begin{tcolorbox}[
    width=\textwidth,
    colback=gray!3,
    colframe=gray!80,
    boxrule=0.5pt,
    arc=2pt,
    left=6pt,
    right=6pt,
    top=6pt,
    bottom=6pt,
    title={\textsc{K-BrowseComp} Example (Original)},
    fonttitle=\bfseries
]
{\scriptsize\ttfamily
\begin{verbatim}
{
  "problem": "한국교육과정평가원에서 2018년 6월 시행한 고3 모의평가 국어 과목 29번에서 다루는 작품에 대하여, 해당 작품의 시인이 생전 열 번째로 간행한 시집에서 열 세 번째로 실려있는 시의 제목은?",
  "answer": "위로 솟구치는 꽃들",
  "type": "multi-hop",
  "category": "문학/도서/언어",
  "expected_chain": [
    {"step": 1, "description": "2018년 6월 시행한 고3 모의평가 국어 과목 29번에서 다루는 작품 확인",
      "sources": ["https://url1"]},
    {"step": 2, "description": "이수익 시인이 생전 열 번째로 간행한 시집 - 소개글 읽어보기", 
      "sources": ["https://url2"]},
    {"step": 3, "description": "처음으로 사랑을 들었다 시집에서 목차 확인", 
      "sources": ["https://url3"]},
    {"step": 4, "description": "13번째 시 식별"}
  ],
  "checklist": [
    {"step": 1, "slot": "다루는 작품", "value": "이수익-결빙의 아버지"},
    {"step": 2, "slot": "10th_collection", "value": "처음으로 사랑을 들었다"},
    {"step": 4, "slot": "final_answer", "value": "위로 솟구치는 꽃들"}
  ],
  "korean_specific_keywords": ["고3 모의평가", "한국 시"]
}
\end{verbatim}
}
\end{tcolorbox}
\begin{tcolorbox}[
    width=\textwidth,
    colback=gray!3,
    colframe=gray!80,
    boxrule=0.5pt,
    arc=2pt,
    left=6pt,
    right=6pt,
    top=6pt,
    bottom=6pt,
    title={\textsc{K-BrowseComp} Example (English Translation)},
    fonttitle=\bfseries
]
{\scriptsize\ttfamily
\begin{Verbatim}[
    breaklines=true,
    breakanywhere=true,
    breaksymbolleft={},
    breaksymbolright={}
]
{
  "problem": "In Question 29 of the Korean language section of the June 2018 high-school mock CSAT administered by KICE, identify the literary work discussed. For the poet of that work, what is the title of the thirteenth poem in the poetry collection that was published as the poet's tenth collection during his lifetime?",
  "answer": "Flowers Surging Upward",
  "type": "multi-hop",
  "category": "Literature / Books / Language",
  "expected_chain": [
    {"step": 1, "description": "Identify the literary work discussed in Question 29 of the June 2018 high-school mock CSAT Korean section.", 
      "sources": ["https://url1"]},
    {"step": 2, "description": "Find the poetry collection published as Lee Su-ik's tenth collection during his lifetime.", 
      "sources": ["https://url2"]},
    {"step": 3, "description": "Check the table of contents of First Hearing Love.", 
      "sources": ["https://url3"]},
    {"step": 4, "description": "Identify the thirteenth poem in that collection."}
  ],
  "checklist": [
    {"step": 1, "slot": "target_work", "value": "Lee Su-ik, The Frozen Father"},
    {"step": 2, "slot": "tenth_collection", "value": "First Hearing Love"},
    {"step": 4, "slot": "final_answer", "value": "Flowers Surging Upward"}
  ],
  "korean_specific_keywords": ["Korean high-school CSAT", "Korean poetry"]
}
\end{Verbatim}
}
\end{tcolorbox}
\caption{
\textbf{Example contributor submission format used in \textsc{K-BrowseComp-Verified}. }
The top shows the original Korean item and the bottom shows its English translation.
Each item was submitted as a structured JSON object containing the problem statement, gold answer, expected reasoning trajectory, intermediate checklist values, and Korean-specific keywords.
}
\label{app:question_example}
\end{figure*}

\section{Dataset Construction Details}\label{app:data_construction_detail}
\textsc{K-BrowseComp-Verified} consists of \textbf{300} human-written Korean browsing problems.
The problems were written by Korean-speaking contributors, including both researchers and non-researchers, who were instructed to create questions that require web browsing.
Each contributor submitted items in a structured JSON format containing the problem statement, gold answer, expected reasoning chain, source URLs, intermediate checklist values, and Korean-specific keywords. An example submission format is shown in Figure~\ref{app:question_example}.

\subsection{Category-wise Performance Analysis}\label{app:category_breakdown}

Table~\ref{tab:category_results} reports model accuracy across the major categories of \textsc{K-BrowseComp-Verified}.
The categories are ordered by the number of questions.
Performance varies substantially across domains, even for the strongest frontier models.

Across most models, \textit{Entertainment \& Media} and \textit{Sports \& Games} are among the highest-performing categories.
These questions often involve public figures, entertainment programs, music releases, sports records, or widely discussed media content that are relatively well represented on the public web.
By contrast, categories such as \textit{Science, IT \& Academia}, \textit{Products, Brands \& Beauty}, and \textit{Education, Colleges \& Exams} remain difficult for nearly all models.
These questions frequently require linking sparse Korean entities across multiple sources, resolving institutional terminology, or navigating semi-structured local web pages.

The gap between global frontier models and Korean-specialized models is also substantial.
\texttt{GPT-5.5} achieves the strongest overall performance at 45.67\%, while Korean-specialized models such as \texttt{K-EXAONE-236B-A23B} and \texttt{A.X-4.0} remain below 11\%.
This gap persists even in categories that contain strongly Korean-specific cultural or institutional information.
The results suggest that current Korean-specialized LLMs still struggle with long-horizon retrieval, multi-step evidence integration, and constraint tracking in realistic Korean web environments.

Several categories additionally exhibit high variance due to sparse but difficult entity structures.
For example, \textit{History, Culture \& Politics} shows relatively high scores for some models despite its difficulty because the category contains only ten questions and several items involve recoverable public records once the correct evidence chain is identified.
In contrast, \textit{Science, IT \& Academia} consistently remains difficult because many questions require cross-source reasoning over researcher profiles, academic affiliations, institutional announcements, and Korean-language metadata.

\begin{table*}[t]
\centering
\scriptsize
\setlength{\tabcolsep}{4.2pt}
\renewcommand{\arraystretch}{1.12}
\begin{tabular}{lccccccccccc}
\toprule
Model
& C1 & C2 & C3 & C4 & C5 & C6 & C7 & C8 & C9 & C10 & Overall \\
& \scriptsize n=109 & \scriptsize n=48 & \scriptsize n=35 & \scriptsize n=26 & \scriptsize n=20
& \scriptsize n=19 & \scriptsize n=15 & \scriptsize n=14 & \scriptsize n=10 & \scriptsize n=4 & \scriptsize n=300 \\
\midrule
GPT-5.5
& 51.4 & 43.8 & 40.0 & 50.0 & 25.0 & 36.8 & 40.0 & 50.0 & 70.0 & 25.0 & 45.7 \\
GPT-5.4-mini
& 39.4 & 31.2 & 20.0 & 34.6 & 5.0 & 31.6 & 26.7 & 21.4 & 20.0 & 50.0 & 30.7 \\
GLM-5.1
& 32.1 & 35.4 & 22.9 & 42.3 & 10.0 & 26.3 & 20.0 & 28.6 & 60.0 & 25.0 & 30.7 \\
DeepSeek-V4-Pro
& 34.9 & 31.2 & 31.4 & 30.8 & 10.0 & 21.1 & 20.0 & 21.4 & 50.0 & 25.0 & 30.0 \\
Gemma-4-31B-it
& 25.7 & 22.9 & 20.0 & 34.6 & 5.0 & 26.3 & 13.3 & 14.3 & 40.0 & 25.0 & 23.3 \\
Gemini-3.1-Flash-Lite
& 12.8 & 14.6 & 5.7 & 3.8 & 5.0 & 15.8 & 0.0 & 7.1 & 40.0 & 25.0 & 11.3 \\
Qwen3.6-35B-A3B
& 10.1 & 14.6 & 8.6 & 23.1 & 5.0 & 10.5 & 0.0 & 7.1 & 40.0 & 25.0 & 12.0 \\
\midrule
K-EXAONE-236B-A23B
& 11.0 & 8.3 & 8.6 & 15.4 & 5.0 & 10.5 & 0.0 & 7.1 & 40.0 & 0.0 & 10.3 \\
A.X-4.0
& 4.6 & 6.2 & 2.9 & 11.5 & 5.0 & 5.3 & 0.0 & 0.0 & 20.0 & 0.0 & 5.3 \\
HyperCLOVAX-Seed-Think-32b
& 2.8 & 4.2 & 2.9 & 0.0 & 0.0 & 0.0 & 0.0 & 0.0 & 10.0 & 0.0 & 2.3 \\
\bottomrule
\end{tabular}

\vspace{0.35em}
\begin{minipage}{0.98\textwidth}
\footnotesize
\textbf{\textit{Category labels.}}
C1: Entertainment \& Media;
C2: Places \& Regions;
C3: Education, Colleges \& Exams;
C4: Sports \& Games;
C5: Science, IT \& Academia;
C6: Food, Drinks \& Restaurants;
C7: Literature, Books \& Language;
C8: Products, Brands \& Beauty;
C9: History, Culture \& Politics;
C10: Economy \& Public Policy.
\end{minipage}

\caption{\textbf{Accuracy by category on \textsc{K-BrowseComp-Verified}. Categories are ordered from left to right by the number of questions.}}
\label{tab:category_results}
\end{table*}

\section{Failure Modes Details}\label{app:failure_modes_details}

This appendix provides additional details for the trajectory-level failure patterns discussed in Section~\ref{sec:trajectory_failures}. 
The main text identifies three representative patterns: candidate capture, unmerged evidence branches, and misbound evidence chains. 
Here, we describe how these patterns appear in concrete browsing trajectories and provide additional examples in Figure~\ref{fig:trajectory_cases_appendix}. 
The examples show that many failures occur after partial retrieval has already succeeded, when the model fails to preserve candidate, constraint, or role state across the trajectory.

\subsection{Candidate Capture Failure Details}
\label{app:candidate_capture}

Candidate capture occurs when the model commits to a plausible candidate before the upstream constraints have been fully verified. 
After this early commitment, later searches are conducted inside the local evidence space of that candidate. 
The final answer may therefore appear supported by retrieved evidence, but the candidate itself is not licensed by the full question.

Figure~\ref{fig:trajectory_cases_appendix}(b) shows an answer-side variant of this pattern.
The question requires preserving the dependency chain from company to brand, H\&B store, and award product. 
The model searches Olive Young award products before fixing the upstream company and brand. 
This search order leads the trajectory toward a plausible \textit{CLIO/Goodal/Peripera} branch, and the model then collects award products from that branch. 
These products are locally plausible, but they are not supported by the company--brand constraint required by the question. 
The resulting output is therefore a set of plausible products, not the unique gold answer, 어성초 흔적 에센스 패드 (\textit{Abib Heartleaf Spot Pad Calming Touch}).

\section{\textsc{Synthetic} Split Diagnostics}\label{app:synthetic_diagnostics}

We conduct additional diagnostics to characterize the 100 synthetic problems.
The goal is to test whether the synthetic split follows the same distribution as \textsc{K-BrowseComp-Verified}, and to identify which factors explain any observed separation.

We first compare the two splits using question embeddings.
All 400 questions are embedded with \texttt{sentence-transformers/paraphrase-multilingual-MiniLM-L12-v2}, producing 384-dimensional vectors.
We then train a cross-validated domain classifier whose task is to predict whether a question comes from \textsc{K-BrowseComp-Verified} or from the \textsc{Synthetic} split.
This classifier reaches ROC AUC 0.8873 $\pm$ 0.0281.
This means that the two splits can be separated from question text alone.

We also run a maximum mean discrepancy test in PCA-reduced embedding space.
Maximum mean discrepancy measures how far apart two sets of points are as distributions under a kernel.
In this analysis, the two point sets are the verified question embeddings and the \textsc{synthetic} question embeddings.
The permutation test randomly shuffles the split labels and recomputes the MMD statistic many times.
The resulting $p$-value measures how often a random label split produces an MMD value at least as large as the observed one.
We obtain RBF MMD$^2=0.031967$ and $p=0.004975$, which indicates a statistically clear embedding-level shift.

To interpret the source of this separation, we train additional classifiers using only simple metadata.
Question length alone gives ROC AUC 0.7885 $\pm$ 0.0558.
Category alone gives ROC AUC 0.7773 $\pm$ 0.0342.
The multi-hop or parallel label alone gives ROC AUC 0.4250 $\pm$ 0.0197.
Using length, category, and type together gives ROC AUC 0.8277 $\pm$ 0.0275.
These results show that much of the embedding separability comes from longer \textsc{synthetic} questions and category rebalancing.
The reasoning-format label contributes little, consistent with the near-identical multi-hop and parallel proportions across the two splits.

\section{Time-Bounded Human Baseline Details}
\label{app:time_bounded_human_baseline}

\paragraph{Sampling and assignment.}
We sampled 30 items from the verified subset and 30 items from the \textsc{Synthetic} split. Sampling was performed separately within each split. The resulting 60 items were distributed across four native Korean speakers. Each item was assigned to exactly one participant and received one attempt.

\paragraph{Search conditions.}
Participants were allowed to use any conventional search engine and browse any publicly accessible web page. They were not allowed to use ChatGPT, web-enabled LLMs, or other LLM-based assistants. Each attempt had a maximum duration of 30 minutes. Participants could submit a single final answer or leave the item unanswered. An unanswered item at the deadline was recorded as no answer and received a score of zero.

\paragraph{Matched model comparison.}
We compared human performance with the existing \texttt{GPT-5.5} outputs for the same 60 items. These outputs were taken from the full benchmark run and were not regenerated after the human study. This preserves item matching without introducing an additional model rollout.

\paragraph{Results.}
Participants produced 9 correct answers, 7 incorrect answers, and 44 no-answer outcomes. Overall accuracy was 15.00\%. Verified accuracy was 3.33\%, with 1 correct answer among 30 items. \textsc{Synthetic} accuracy was 26.67\%, with 8 correct answers among 30 items. The 44 no-answer attempts consumed the complete 30-minute allowance and corresponded to 22 person-hours of unsuccessful search.

On the same items, \texttt{GPT-5.5} answered 26 of 60 questions correctly, yielding 43.33\% accuracy. It answered 15 of 30 verified items and 11 of 30 \textsc{Synthetic} items correctly. These results provide a time-bounded calibration of discovery difficulty. They do not measure human performance with unrestricted time, collaboration, specialized tools, or access to the gold evidence path.

\paragraph{Interpretation.}
The low human accuracy on the verified sample indicates that independently discovering the intended evidence path can require substantial search effort even for native Korean speakers. It does not imply that the gold answers are unverifiable. Every gold answer and trajectory was separately checked against public evidence during dataset validation.

The higher human accuracy on the \textsc{Synthetic} sample is consistent with the split being optimized for model failure. It also cautions against interpreting the \textsc{Synthetic} split as universally harder for humans. The sample is too small to establish a precise split-level ranking of human difficulty.

\section{Trajectory-Level Failure Diagnostics for Korean Open-Weight Models}\label{app:korean_model_trajectory_analysis}

This appendix expands the analysis of Korean open-weight models in Section~\ref{sec:trajectory_failures}.
The main results show that their low performance does not come from a single failure type.
Instead, the trajectories reveal different bottlenecks in the browsing loop: some models fail to form a useful candidate state after retrieval, while others retrieve relevant evidence but lose the dependency chain needed for later search and final answer selection.

\paragraph{Shallow evidence control in \texttt{A.X-4.0}.}
Figure~\ref{fig:ax_weak_evidence_control} illustrates a failure in which \texttt{A.X-4.0} reaches relevant Korean historical evidence, but does not convert the retrieved snippets into a maintained candidate ledger.
The question requires one historical person to satisfy several constraints at once: civil-service examination records, surname-and-clan evidence linked to a modern athlete, inspection-related office history, and a memorial academy in the person's hometown.
The model issues a single broad query that concatenates most of these clues.
This search surfaces locally relevant names, including Jeong Hwang, Min Yeong-hwan, and Kim Jong-seo, but the model evaluates them as separate snippet-level candidates.
The same-origin modern-athlete clue is never used as a binding constraint, so the trajectory does not keep track of which historical candidates remain valid after each check.
The final \textit{I don't know} response therefore occurs after partial retrieval has succeeded.
The bottleneck is state formation: the model sees relevant evidence, but fails to organize it into a shared candidate table that can support elimination and verification.

\paragraph{Cross-source chain drift in \texttt{K-EXAONE-236B-A23B}.}
Figure~\ref{fig:exaone_cross_source_chain_drift} shows a later-stage failure.
Unlike the \texttt{A.X-4.0} case, \texttt{K-EXAONE-236B-A23B} initially reaches the right web region.
Its first search retrieves the KakaoTalk free-emoticon event and a plausible list of distributed character emoticons.
At this point, the model should freeze the event-side state, identify the smaller-animal character, and then search for the creator and official YouTube channel metadata.
The trajectory instead moves to creator search before stabilizing the target emoticon.
Subsequent queries contain misspelled character names, malformed creator strings, and unrelated YouTube terms.
Although the model later finds a plausible creator-like entity and a channel date, it cannot verify the event--emoticon--creator--channel link.
This failure shows that successful initial retrieval is not enough.
The intermediate entity must remain stable across source boundaries; otherwise, later searches may be locally plausible but no longer licensed by the original question.

\paragraph{Trajectory completion and protocol stability.}
The remaining Korean open-weight models show different bottlenecks.
\texttt{HyperCLOVAX-SEED-Think-32B} often begins a reasonable browsing process, but its trajectories do not reliably converge to an exact final answer.
These failures are not limited to one content type.
They include unresolved constraint checks, unstable candidate states, and final responses that remain uncertain or under-specified.
\texttt{Kanana-2-30B-A3B-Thinking-2601} shows an earlier bottleneck in the evaluation loop.
Because it frequently fails to emit valid tool-call objects under our browsing harness, many runs do not become completed search trajectories.
This indicates that tool-use protocol reliability remains a prerequisite for evaluating higher-level browsing skills such as evidence selection and cross-source reasoning.

\paragraph{Summary.}
These examples clarify why Korean open-weight models score poorly on \textsc{K-BrowseComp}.
The issue is not only whether a model can retrieve Korean web pages.
A successful browsing agent must convert retrieved pages into a persistent trajectory state: candidate ledgers, constraint checks, entity-role bindings, and final-answer commitments.
The Korean open-weight models fail at different points in this process.
\texttt{A.X-4.0} often stops at shallow snippet-level evidence, \texttt{K-EXAONE-236B-A23B} loses dependency chains after relevant retrieval, \texttt{HyperCLOVAX-SEED-Think-32B} struggles to stabilize final answers, and \texttt{Kanana-2-30B-A3B-Thinking-2601} is limited by tool-call protocol failures.

\section{\textsc{Synthetic} Split Trajectory Diagnostics}
\label{app:synthetic_trajectory_diagnostics}

This appendix expands the trajectory-level analysis of the \textsc{synthetic} split.
The \textsc{synthetic} split is constructed to target failure modes that can be instantiated reliably by the generation pipeline, with semi-structured parsing (F4) and constraint tracking (F7) appearing most frequently among accepted items.
The examples in Figures~\ref{fig:synthetic_filesize_failure} and~\ref{fig:synthetic_kbo_constraint_failure} show how these targeted modes also appear in completed model trajectories.
In both cases, the model does not fail because it searches in an unrelated region of the web.
It reaches the correct source family or recovers the main intermediate entity, but loses the page-level or candidate-level state needed to produce the exact final answer.

\paragraph{Metadata-field extraction after successful source localization.}
Figure~\ref{fig:synthetic_filesize_failure} illustrates a semi-structured parsing failure in a repository metadata question.
The question asks for the file size displayed for a PDF attached to a KOPRI repository record.
The required state is narrow: the solver must preserve the target repository item and read the file-size field attached to that specific PDF.
The model first retrieves the correct topical context, namely the 2015 KOPRI press release about the Araon Antarctic mid-ocean ridge study.
It also reaches the relevant repository neighborhood.
However, the trajectory does not keep the target item fixed when reading the file metadata.
Because the search results contain nearby PDF records with similar filenames and related institutional context, the model selects a neighboring file-size value, 1.56 MB, while the gold value is 698.85 kB.

This case shows a page-level state failure.
The model has enough evidence to enter the right source family, but it does not bind the final extraction step to the correct repository record.
The error is therefore not a failure of broad topical retrieval.
It is a failure to preserve an exact source pointer and extract the requested field from that source.
This is the kind of operation that appears simple after the target page is known, but remains difficult for browsing agents because visually or textually adjacent records create plausible distractors.

\paragraph{Record-level ledger failure after entity recovery.}
Figure~\ref{fig:synthetic_kbo_constraint_failure} shows a constraint-tracking failure in a KBO record question.
The first part of the question identifies the target player through 2022 pitching clues: a Kiwoom Heroes right-handed pitcher who led the league in ERA, strikeouts, and innings.
The model correctly recovers 안우진 (\textit{An Woo-jin}) as the target player.
It then searches for his 2026 appearance records and reaches relevant baseball-record pages and reports about specific appearances.
The remaining task is not entity identification, but comparison over a stable game-level ledger.
The solver must maintain rows containing the date, appearance, and opponent AVG values for the player's 2026 games, then select the unique maximum.

The model does not maintain this ledger.
Instead, it selects a plausible later appearance date from the retrieved evidence without verifying the maximum opponent AVG condition across all appearances.
Its final answer is 05.14, while the gold answer is 05.08.
This demonstrates a candidate-level state failure.
The model has already found the central entity and entered the right record space, but the final comparison requires a structured table over all relevant games.
Without such a table, the trajectory can remain locally plausible while failing the global constraint.

\paragraph{Relation to the \textsc{synthetic} generation pipeline.}
These two cases clarify what the \textsc{synthetic} split contributes beyond the human-written verified set.
The generation pipeline does not merely produce questions with obscure answers.
It creates items in which the answer is easy to verify once the correct state is fixed, but difficult to obtain unless the model preserves that state across the trajectory.
For F4-style items, the key state is the exact page item and field to be read.
For F7-style items, the key state is a candidate or record ledger that must survive later filtering and comparison.
The accepted \textsc{synthetic} questions therefore stress operations that are common in real browsing: reading institutional metadata, selecting among neighboring records, maintaining candidate tables, and enforcing final constraints after the main entity has already been found.

\paragraph{Takeaway.}
The \textsc{synthetic} split should be interpreted as a complementary diagnostic stress split, not as a replacement for \textsc{K-BrowseComp-Verified}.
Its category and surface profile differ from the verified set, but its trajectory failures reveal the same broader weakness: models often retrieve relevant Korean web evidence without converting it into a persistent state for exact extraction or final selection.
Figures~\ref{fig:synthetic_filesize_failure} and~\ref{fig:synthetic_kbo_constraint_failure} show this at two granularities.
The repository example fails at the source-field level, while the KBO example fails at the candidate-ledger level.
Together, they explain why accuracy on the \textsc{synthetic} split remains low even when models reach relevant sources.

\section{AI assistants in research/writing}

In \autoref{sec:K-BrowseComp-llm}, we used claude-code to generate the \textsc{Synthetic} subset of \textsc{K-BrowseComp} and in \autoref{sec:experimental_setting}, we include multiple LLMs as our baselines in \textsc{K-BrowseComp}. For writing, we mainly used AI assistants for revising grammar and fluency.

\section{Computational experiments}

\paragraph{Model Size} Among the open-weight baselines we evaluate, several disclose their parameter counts. Three are dense models: Gemma-4-31B-it (31B parameters), HCX-SEED-Think-32B (32B parameters), and A.X-4.0 (72B parameters; continual pre-training of Qwen2.5-72B on Korean data). Five are Mixture-of-Experts (MoE) models, for which we report total / active parameter counts: Qwen3.6-35B-A3B (35B / 3B), Kanana-2-30B-A3B-Think (30B / 3B), K-EXAONE-236B-A23B (236B / 23B), GLM-5.1 (744B / 40B), and DeepSeek-V4-Pro (1.6T / 49B). For the closed-weight baselines (GPT-5.5, GPT-5.4-mini, and Gemini-3.1-Flash-Lite), parameter counts are not publicly disclosed by the providers.

\paragraph{Compute Budget} All experiments ran on openrouter API calls except for Korean models that were not supported on the platform. The total API cost for this project was approximately 500 USD.

\begin{figure*}[p]
\centering
{\fontsize{6.65}{7.35}\selectfont
\setlength{\fboxsep}{0pt}
\setlength{\parindent}{0pt}

\begin{failurecase}{a}{F1}{Ineffective initial search direction}

\begin{trajblock}{kbBlue}{\kbBadge{kbBlue}{Q}\hspace{3pt} Question}
다음 조건을 모두 만족하는 말은?
가족과 함께 일을 하는 기수와 경기를 뛴 적이 있고, 한 신문의 편집부장에게 경기 전 모습으로 기대를 받았으며, 가장 오래 보유한 마주는 2013년 11월에 처음 등록했고, 단 한 번의 우승 기록을 보유한다.

\eng{Which horse satisfies all of the following conditions: it has raced with a jockey who worked together with family members, was expected to perform well before a race based on its appearance according to a newspaper editor-in-chief, has an owner who was first registered in November 2013 and remained the horse’s longest-associated owner, and has exactly one career win?}

\vspace{1pt}
\textcolor{kbBlue!80!black}{\textbf{Gold:}} 퍼플망고
\eng{Purple Mango}
\end{trajblock}

\begin{trajblock}{kbState}{\kbBadge{kbState}{S}\hspace{3pt} Required state}
Form a racehorse candidate set before searching ownership and win-record constraints.
\end{trajblock}

\begin{trajblock}{kbGreen}{\kbBadge{kbGreen}{G}\hspace{3pt} Gold trajectory}
\gstep{1} Identify the jockey clue.\\
\gstep{2} Find the newspaper editor's pre-race comment.\\
\gstep{3} Verify the longest-owning owner's registration date.\\
\gstep{4} Check that the horse has exactly one win.\\
\textcolor{kbGreen!75!black}{$\rightarrow$ \textbf{퍼플망고} \eng{Purple Mango}}
\end{trajblock}

\begin{trajblock}{kbRed}{\kbBadge{kbRed}{M}\hspace{3pt} Model trajectory}
\textit{Raw search queries}
\searchquery{"편집부장" "경기 전" "말" "기수"

\eng{"editor-in-chief" "before the race" horse jockey}}
\searchquery{"편집부장" "예시장" "경마 말"

\eng{"editor-in-chief" "paddock" "horse racing"}
}
\searchquery{"2013년 11월" "마주" "등록" 경주마

\eng{"November 2013" horse owner registration racehorse}
}

\rstep{1} Searches generic editor/race/horse terms.\\
\rstep{2} Drifts into general paddock or racing-advice pages.\\
\rstep{3} Never forms a concrete racehorse candidate set.\\
\rstep{4} \textcolor{kbRed!80!black}{\textbf{Result: I don't know}}
\end{trajblock}

\begin{trajblock}{kbOrange}{\kbBadge{kbOrange}{!}\hspace{3pt} Failure point}
The trajectory breaks at the first retrieval step. The initial queries are too broad to surface candidate horses, so the later ownership and win-record checks have no entity to attach to.
\end{trajblock}
\raggedright\textit{\textcolor{black!55}{Example model behavior: GPT-5.5}}
\end{failurecase}
\hfill
\begin{failurecase}{b}{F2}{Search-access structure failure}

\begin{trajblock}{kbBlue}{\kbBadge{kbBlue}{Q}\hspace{3pt} Question}
KSIC 코드 J63391로 분류되며, 상장폐지 이력이 있고, 2026년 5월 기준 중소기업이며, 본사가 강남구에 있는 회사의 공식 유튜브 채널에서 업로드 일자를 오래된 순으로 정렬했을 때 두 번째 영상 제목은?

\eng{What is the title of the second-oldest video on the official YouTube channel of a company that is classified under KSIC code J63391, has a delisting history, is categorized as an SME as of May 2026, and is headquartered in Gangnam-gu?}

\vspace{1pt}
\textcolor{kbBlue!80!black}{\textbf{Gold:}} MBC드라마 `마의' OST Part.1\_소향\_오직 단 하나\_소향 인터뷰

\eng{MBC Drama \texttt{Horse Doctor} OST Part 1\_Sohyang\_Only One\_Sohyang Interview}
\end{trajblock}

\begin{trajblock}{kbState}{\kbBadge{kbState}{S}\hspace{3pt} Required state}
Access a channel-level ordered list, then select the second-oldest video.
\end{trajblock}

\begin{trajblock}{kbGreen}{\kbBadge{kbGreen}{G}\hspace{3pt} Gold trajectory}
\gstep{1} Identify the company under KSIC J63391.\\
\gstep{2} Verify delisting history, SME status, and Gangnam headquarters.\\
\gstep{3} Open the official YouTube channel.\\
\gstep{4} Sort videos from oldest to newest.\\
\gstep{5} Read the second video title.\\
\textcolor{kbGreen!75!black}{$\rightarrow$ \textbf{MBC드라마 `마의' OST Part.1\_소향\_오직 단 하나\_소향 인터뷰}}

\eng{MBC Drama \texttt{Horse Doctor} OST Part 1\_Sohyang\_Only One\_Sohyang Interview}
\end{trajblock}

\begin{trajblock}{kbRed}{\kbBadge{kbRed}{M}\hspace{3pt} Model trajectory}
\textit{Raw search queries}
\searchquery{KSIC J63391 한국표준산업분류 J63391

\eng{KSIC J63391 Korean Standard Industrial Classification}}
\searchquery{"J63391" 회사 상장폐지 강남구 중소기업

\eng{"J63391" company delisting Gangnam-gu SME}}
\searchquery{"63391" "데이터베이스" "온라인"

\eng{"63391" database online}}

\rstep{1} Searches company-identification evidence.\\
\rstep{2} Does not reproduce the YouTube channel ordering operation.\\
\rstep{3} Picks a plausible company/video result from search.\\
\rstep{4} \textcolor{kbRed!80!black}{\textbf{Result: 에프앤가이드 온라인 기업설명회}}
\eng{FnGuide Online Corporate Briefing}
\end{trajblock}

\begin{trajblock}{kbOrange}{\kbBadge{kbOrange}{!}\hspace{3pt} Failure point}
The answer is hidden behind an access structure: official channel navigation and oldest-first ordering. General web search surfaces plausible videos, but not the required ordered second item.
\end{trajblock}
\raggedright\textit{\textcolor{black!55}{Example model behavior: GPT-5.5}}
\end{failurecase}
}

\vspace{3pt}
\caption{
\textbf{Representative search-direction and access-structure failures in \textsc{K-BrowseComp}.}
Each panel shows the question, gold answer, required trajectory state, intended gold trajectory, model search queries, and the point where the trajectory first diverges.
F1 illustrates an ineffective initial search direction, where broad generic queries fail to surface any concrete racehorse candidates and leave later constraints unattached.
F2 illustrates a search-access structure failure, where the required answer depends on reproducing an internal platform operation, namely oldest-first video ordering within an official YouTube channel.
}
\label{fig:failure_modes_appendix_f1_f2}
\end{figure*}

\begin{figure*}[p]
\centering
{\fontsize{6.65}{7.35}\selectfont
\setlength{\fboxsep}{0pt}
\setlength{\parindent}{0pt}

\begin{failurecase}{a}{F3}{Cross-source hopping failure}
\begin{trajblock}{kbBlue}{\kbBadge{kbBlue}{Q}\hspace{3pt} Question}
퍼듀대학교 공식 한인회 웹사이트를 개발한 초대 멤버 중, 석박통합과정으로 서울대학교에 입학한 사람이 서울대학교에서 소속된 학과 이름은?

\eng{Among the founding members who developed the official Purdue Korean Association website, what was the name of the department affiliated with the person who entered Seoul National University through an integrated M.S./Ph.D. program?}

\vspace{1pt}
\textcolor{kbBlue!80!black}{\textbf{Gold:}} 협동과정 인공지능 전공
\eng{Interdisciplinary Program in Artificial Intelligence}
\end{trajblock}

\begin{trajblock}{kbState}{\kbBadge{kbState}{S}\hspace{3pt} Required state}
Preserve the chain:
PKA website developer $\rightarrow$ founding member $\rightarrow$ SNU integrated Ph.D. entrant $\rightarrow$ Korean department name.
\end{trajblock}

\begin{trajblock}{kbGreen}{\kbBadge{kbGreen}{G}\hspace{3pt} Gold trajectory}
\gstep{1} Locate Purdue Korean Association website/member evidence.\\
\gstep{2} Identify the founding web developer.\\
\gstep{3} Connect the person to SNU integrated graduate admission.\\
\gstep{4} Read the Korean program affiliation.\\
\textcolor{kbGreen!75!black}{$\rightarrow$ \textbf{협동과정 인공지능 전공}}
\eng{Interdisciplinary Program in Artificial Intelligence}
\end{trajblock}

\begin{trajblock}{kbRed}{\kbBadge{kbRed}{M}\hspace{3pt} Model trajectory}
\textit{Raw search queries}
\searchquery{퍼듀대학교 공식 한인회 웹사이트 개발 초대 멤버 석박통합과정 서울대학교

\eng{Purdue Korean Association website founding developer integrated M.S./Ph.D. Seoul National University}}
\searchquery{"Purdue Korean Association" "website" "Seoul National University"}
\searchquery{site:purdueka.org "서울대학교" OR "SNU" "석박통합"

\eng{site:purdueka.org "Seoul National University" OR "SNU" integrated Ph.D.}}
\searchquery{"purdueka.org" "Developed" "member" "SNU"}

\rstep{1} Reaches relevant Purdue/SNU search regions.\\
\rstep{2} Does not bind the PKA developer to the SNU page.\\
\rstep{3} Normalizes the affiliation to a more generic department.\\
\rstep{4} \textcolor{kbRed!80!black}{\textbf{Result: 컴퓨터공학부}}
\eng{Department of Computer Science and Engineering}
\end{trajblock}

\begin{trajblock}{kbOrange}{\kbBadge{kbOrange}{!}\hspace{3pt} Failure point}
The trajectory loses the cross-source link between the Purdue-side person and the SNU-side program. The final answer is plausible in the SNU context, but not the department attached to the target person.
\end{trajblock}
\raggedright\textit{\textcolor{black!55}{Example model behavior: GPT-5.5}}
\end{failurecase}
\hfill
\begin{failurecase}{b}{F4}{Semi-structured parsing failure}

\begin{trajblock}{kbBlue}{\kbBadge{kbBlue}{Q}\hspace{3pt} Question}
정가가 17,000원 이상이고 시/에세이 분류에 해당하며, 성별이 다른 저자 2명이 있고, 종이책 ISBN 맨 뒤 네 자릿수의 합이 19인 책의 목차에서 가장 긴 챕터 제목은?

\eng{What is the longest chapter title in the table of contents of a book that satisfies all of the following conditions: the list price is at least 17,000 KRW, the category is poetry/essay, the book has two authors of different genders, and the sum of the last four digits of the print ISBN is 19?}

\vspace{1pt}
\textcolor{kbBlue!80!black}{\textbf{Gold:}} 착한 여자는 천국에 가고 나쁜 여자는 어디에나 가지만 어리석은 여자는 군부대로 강연을 간다

\eng{Good women go to heaven, bad women go everywhere, but foolish women go to military bases to give lectures}
\end{trajblock}

\begin{trajblock}{kbState}{\kbBadge{kbState}{S}\hspace{3pt} Required state}
Extract and join book-page fields: price, category, authors, ISBN, and table of contents.
\end{trajblock}

\begin{trajblock}{kbGreen}{\kbBadge{kbGreen}{G}\hspace{3pt} Gold trajectory}
\gstep{1} Filter books by category and price.\\
\gstep{2} Verify two authors with different genders.\\
\gstep{3} Sum the last four digits of the print ISBN.\\
\gstep{4} Parse the table of contents.\\
\gstep{5} Select the longest chapter title.
\end{trajblock}

\begin{trajblock}{kbRed}{\kbBadge{kbRed}{M}\hspace{3pt} Model trajectory}
\textit{Raw search queries}
\searchquery{"종이책 ISBN" "시/에세이" "목차" "17,000원" "저자"

\eng{"print ISBN" "poetry/essay" "table of contents" "17,000 KRW" author}}
\searchquery{site:aladin.co.kr/shop/wproduct.aspx "시/에세이" "정가" "ISBN" "목차" "17,000원" "지은이" "옮긴이"

\eng{site:aladin.co.kr/shop/wproduct.aspx "poetry/essay" price ISBN "table of contents" "17,000 KRW" author translator}}
\searchquery{"시/에세이" "17,000원" "ISBN" "목차" "지음" "공저"

\eng{"poetry/essay" "17,000 KRW" ISBN "table of contents" co-author}}

\rstep{1} Searches the correct book domains.\\
\rstep{2} Encounters many partially matching book pages.\\
\rstep{3} Does not extract all required fields from one page.\\
\rstep{4} \textcolor{kbRed!80!black}{\textbf{Result: I don't know}}
\end{trajblock}

\begin{trajblock}{kbOrange}{\kbBadge{kbOrange}{!}\hspace{3pt} Failure point}
The failure begins after domain selection. The model reaches semi-structured book pages, but cannot reliably parse and combine the relevant slots into one candidate.
\end{trajblock}
\raggedright\textit{\textcolor{black!55}{Example model behavior: GPT-5.5}}
\end{failurecase}
}

\vspace{3pt}
\caption{
\textbf{Representative cross-source linking and semi-structured parsing failures in \textsc{K-BrowseComp}.}
Each panel shows the question, gold answer, required trajectory state, intended gold trajectory, model search queries, and the point where the trajectory first diverges.
F3 illustrates a cross-source hopping failure, where the model reaches both the Purdue-side and SNU-side evidence regions but fails to preserve the person-level link across sources.
F4 illustrates a semi-structured parsing failure, where the model reaches relevant book pages but fails to extract and combine price, category, author, ISBN, and table-of-contents fields from the same candidate.
}
\label{fig:failure_modes_appendix_f3_f4}
\end{figure*}

\begin{figure*}[p]
\centering
{\fontsize{6.65}{7.35}\selectfont
\setlength{\fboxsep}{0pt}
\setlength{\parindent}{0pt}

\begin{failurecase}{a}{F5}{Search-result selection failure}

\begin{trajblock}{kbBlue}{\kbBadge{kbBlue}{Q}\hspace{3pt} Question}
엔데믹 이후 신규 개통된 철도 노선 중 2026년에 연결되는 두 역을 찾고, 2025년 기준 이용객이 더 적은 역을 고른다. 그 역과 가장 가까운 산, 같은 지역 내 높이가 가장 비슷한 다른 산, 그 산 이름의 유래 동물을 확인하여 2026년 이후 가장 먼저 돌아오는 해당 동물의 해를 답하라.

\eng{Among railway lines newly opened after the endemic period, identify the two stations connected in 2026 and select the station with lower ridership as of 2025. Then find the mountain closest to that station, identify another mountain in the same region with the most similar elevation, determine the animal associated with the origin of that mountain's name, and answer the first year after 2026 corresponding to that animal in the zodiac cycle.}

\vspace{1pt}
\textcolor{kbBlue!80!black}{\textbf{Gold:}} 2033년
\eng{2033}
\end{trajblock}

\begin{trajblock}{kbState}{\kbBadge{kbState}{S}\hspace{3pt} Required state}
Select the correct station branch before continuing to mountain, animal-origin, and zodiac-year steps.
\end{trajblock}

\begin{trajblock}{kbGreen}{\kbBadge{kbGreen}{G}\hspace{3pt} Gold trajectory}
\gstep{1} Identify the newly opened line and two stations connected in 2026.\\
\gstep{2} Select the lower-ridership station as of 2025.\\
\gstep{3} Find the nearest mountain.\\
\gstep{4} Find the closest-height mountain in the same region.\\
\gstep{5} Resolve the animal origin and next year.\\
\textcolor{kbGreen!75!black}{$\rightarrow$ \textbf{2033년}}
\eng{2033}
\end{trajblock}

\begin{trajblock}{kbRed}{\kbBadge{kbRed}{M}\hspace{3pt} Model trajectory}
\textit{Raw search queries}
\searchquery{2026년 연결되는 두 역 신규 개통 철도 노선 2024 개통 역 이용객 2025

\eng{newly opened railway line stations connected in 2026 ridership 2025}}
\searchquery{서화성역 가장 가까운 산

\eng{nearest mountain to Seo-Hwaseong Station}}
\searchquery{서화성역 주변 산 고도 화성시 남양읍

\eng{mountain elevation near Seo-Hwaseong Station Hwaseong Namyang-eup}}

\rstep{1} Retrieves station-level evidence.\\
\rstep{2} Moves into the Seo-Hwaseong station branch.\\
\rstep{3} Selects the wrong downstream mountain/animal chain.\\
\rstep{4} \textcolor{kbRed!80!black}{\textbf{Result: 2036년}}
\eng{2036}
\end{trajblock}

\begin{trajblock}{kbOrange}{\kbBadge{kbOrange}{!}\hspace{3pt} Failure point}
The model finds relevant station evidence, but chooses the wrong downstream candidate from nearby search results. The final year follows a coherent but incorrect entity chain.
\end{trajblock}
\raggedright\textit{\textcolor{black!55}{Example model behavior: GPT-5.5}}
\end{failurecase}
\hfill
\begin{failurecase}{b}{F6}{Sparse entity normalization failure}

\begin{trajblock}{kbBlue}{\kbBadge{kbBlue}{Q}\hspace{3pt} Question}
중앙대학교 명예교수이며, 법인설립허가 내부결재일이 1956년인 문화체육관광부 소관의 사단법인 회장을 역임했고, YTN 라디오 등에서 4차 산업혁명과 미래 교육 관련 인터뷰 및 강연을 한 인물은?

\eng{Which person satisfies all of the following conditions: they are an emeritus professor at Chung-Ang University, served as president of an incorporated association under the Ministry of Culture, Sports and Tourism whose internal approval date for establishment permission was in 1956, and gave interviews and lectures on the Fourth Industrial Revolution and future education on YTN Radio and related media?}

\vspace{1pt}
\textcolor{kbBlue!80!black}{\textbf{Gold:}} 남영준
\eng{Nam Young-jun}
\end{trajblock}

\begin{trajblock}{kbState}{\kbBadge{kbState}{S}\hspace{3pt} Required state}
Resolve a sparse Korean person entity across university, association, and media-interview clues.
\end{trajblock}

\begin{trajblock}{kbGreen}{\kbBadge{kbGreen}{G}\hspace{3pt} Gold trajectory}
\gstep{1} Verify the Chung-Ang emeritus-professor clue.\\
\gstep{2} Connect the association to Korea Library Association.\\
\gstep{3} Check chair history and 1956 incorporation clue.\\
\gstep{4} Match YTN/interview topics to the same person.\\
\textcolor{kbGreen!75!black}{$\rightarrow$ \textbf{남영준}}
\eng{Nam Young-jun}
\end{trajblock}

\begin{trajblock}{kbRed}{\kbBadge{kbRed}{M}\hspace{3pt} Model trajectory}
\textit{Raw search queries}
\searchquery{중앙대학교 명예교수 "4차 산업혁명" "미래 교육" "YTN 라디오"

\eng{Chung-Ang University emeritus professor "Fourth Industrial Revolution" "future education" "YTN Radio"}}
\searchquery{"이성규" "중앙대학교" "명예교수" "4차 산업혁명"

\eng{"Lee Seong-gyu" "Chung-Ang University" emeritus professor "Fourth Industrial Revolution"}}
\searchquery{site:mcst.go.kr 사단법인 법인설립허가 1956 "한국도서관협회" 회장 중앙대학교 명예교수

\eng{site:mcst.go.kr incorporated association establishment approval 1956 "Korea Library Association" president Chung-Ang University emeritus professor}}

\rstep{1} Enters the correct institution and topic area.\\
\rstep{2} Tests nearby scholar names.\\
\rstep{3} Confuses a related library-science figure.\\
\rstep{4} \textcolor{kbRed!80!black}{\textbf{Result: 남태우}}
\eng{Nam Tae-woo}
\end{trajblock}

\begin{trajblock}{kbOrange}{\kbBadge{kbOrange}{!}\hspace{3pt} Failure point}
The model stays in the right disciplinary neighborhood but normalizes the clues to a nearby person. The final name is locally plausible, yet it does not satisfy the full entity chain.
\end{trajblock}
\raggedright\textit{\textcolor{black!55}{Example model behavior: GPT-5.5}}
\end{failurecase}
}

\vspace{3pt}
\caption{
\textbf{Representative search-result selection and entity-normalization failures in \textsc{K-BrowseComp}.}
Each panel shows the question, gold answer, required trajectory state, intended gold trajectory, model search queries, and the point where the trajectory first diverges.
F5 illustrates a search-result selection failure, where the model retrieves relevant station evidence but follows the wrong downstream mountain and zodiac branch.
F6 illustrates a sparse entity normalization failure, where the model stays in the correct disciplinary neighborhood but maps the clues to a nearby person instead of the target entity.
}
\label{fig:failure_modes_appendix_f5_f6}
\end{figure*}

\begin{figure*}[p]
\centering
{\fontsize{6.65}{7.35}\selectfont
\setlength{\fboxsep}{0pt}
\setlength{\parindent}{0pt}

\begin{failurecase}{a}{F7}{Constraint-tracking failure}

\begin{trajblock}{kbBlue}{\kbBadge{kbBlue}{Q}\hspace{3pt} Question}
골든디스크어워즈 디지털 음원 부문에서 3곡 이상 수상했고, 발매곡 가사에 멤버 이름이 들어가며, 멤버 중 한 명이 고등학교 배경 드라마에 출연했고, 멤버 국적이 2개 이상인 K-pop 그룹은?

\eng{Which K-pop group satisfies all of the following conditions: the group won awards for at least three songs in the Digital Song Division of the Golden Disc Awards, has a released song whose lyrics contain the name of one of the members, has at least one member who appeared in a drama set in a high school, and consists of members from at least two different nationalities?}

\vspace{1pt}
\textcolor{kbBlue!80!black}{\textbf{Gold:}} 미쓰에이
\eng{miss A}
\end{trajblock}

\begin{trajblock}{kbState}{\kbBadge{kbState}{S}\hspace{3pt} Required state}
Maintain one candidate ledger across all parallel constraints.
\end{trajblock}

\begin{trajblock}{kbGreen}{\kbBadge{kbGreen}{G}\hspace{3pt} Gold state}
Golden Disc digital-song winners with $\geq 3$ awarded songs\\
\hspace*{1em}$\cap$ lyrics mentioning a member name\\
\hspace*{1em}$\cap$ member appeared in a high-school drama\\
\hspace*{1em}$\cap$ multinational membership\\
\textcolor{kbGreen!75!black}{$\rightarrow$ \textbf{미쓰에이}}
\eng{miss A}
\end{trajblock}

\begin{trajblock}{kbRed}{\kbBadge{kbRed}{M}\hspace{3pt} Model trajectory}
\textit{Raw search queries}
\searchquery{Golden Disc Awards Digital Song Bonsang winners groups 3 songs TWICE BLACKPINK}
\searchquery{(G)I-DLE Golden Disc Awards Digital Song Bonsang Tomboy Queencard Fate}
\searchquery{(G)I-DLE My Bag lyrics Miyeon Minnie Soyeon Yuqi Shuhua names}

\rstep{1} Finds a strong partial candidate: (G)I-DLE.\\
\rstep{2} Verifies award-history evidence for that candidate.\\
\rstep{3} Searches lyric evidence inside the same branch.\\
\rstep{4} Does not enforce all remaining constraints.\\
\rstep{5} \textcolor{kbRed!80!black}{\textbf{Result: (여자)아이들}}
\eng{(G)I-DLE}
\end{trajblock}

\begin{trajblock}{kbOrange}{\kbBadge{kbOrange}{!}\hspace{3pt} Failure point}
The model commits to a candidate that satisfies salient early constraints. Later searches support that candidate locally, but the full constraint intersection is never checked.
\end{trajblock}
\raggedright\textit{\textcolor{black!55}{Example model behavior: GPT-5.5}}
\end{failurecase}
\hfill
\begin{failurecase}{b}{F8}{Intermediate reasoning failure}

\begin{trajblock}{kbBlue}{\kbBadge{kbBlue}{Q}\hspace{3pt} Question}
2024년 6월 네이버 블로그 챌린지의 최대 포인트 $X$, 2023년 5월 챌린지에서 레벨 2 달성 시 포인트를 받은 인원 $Y$, 2022년 6월 챌린지 4주 참여 포인트 $Z$를 더한 수의 약수 개수는?

\eng{What is the number of divisors of the sum of the following three quantities: the maximum point reward $X$ in the June 2024 Naver Blog Challenge, the number of users $Y$ who received points for reaching Level 2 in the May 2023 challenge, and the four-week participation reward $Z$ in the June 2022 challenge?}

\vspace{1pt}
\textcolor{kbBlue!80!black}{\textbf{Gold:}} 60개
\eng{60}
\end{trajblock}

\begin{trajblock}{kbState}{\kbBadge{kbState}{S}\hspace{3pt} Required state}
Retrieve $X$, $Y$, and $Z$, then compute the divisor count of $X+Y+Z$.
\end{trajblock}

\begin{trajblock}{kbGreen}{\kbBadge{kbGreen}{G}\hspace{3pt} Gold trajectory}
\gstep{1} Find the 2024 June challenge reward value $X$.\\
\gstep{2} Find the 2023 May level-2 recipient count $Y$.\\
\gstep{3} Find the 2022 June four-week reward value $Z$.\\
\gstep{4} Compute $X+Y+Z$.\\
\gstep{5} Count the divisors.\\
\textcolor{kbGreen!75!black}{$\rightarrow$ \textbf{60개}}
\eng{60}
\end{trajblock}

\begin{trajblock}{kbRed}{\kbBadge{kbRed}{M}\hspace{3pt} Model trajectory}
\textit{Raw search queries}
\searchquery{2024년 6월 네이버 블로그 챌린지 첫 주부터 참여 3주 연속 성공 최대 포인트

\eng{June 2024 Naver Blog Challenge maximum reward for three consecutive successful weeks}}
\searchquery{포토덤프 챌린지 3주 연속 성공 네이버페이 포인트 최대 2024

\eng{Photo Dump Challenge maximum Naver Pay reward for three consecutive successful weeks 2024}}
\searchquery{2023년 5월 네이버 블로그 챌린지 레벨 2 달성 2천원 포인트 몇명

\eng{May 2023 Naver Blog Challenge number of Level-2 users receiving 2,000 KRW points}}

\rstep{1} Retrieves the right kind of slot values.\\
\rstep{2} Continues searching year-specific challenge evidence.\\
\rstep{3} Miscomputes the final arithmetic or divisor count.\\
\rstep{4} \textcolor{kbRed!80!black}{\textbf{Result: 96}}
\end{trajblock}

\begin{trajblock}{kbOrange}{\kbBadge{kbOrange}{!}\hspace{3pt} Failure point}
The failure starts after retrieval. The model fills relevant intermediate slots, but the final numerical operation produces the wrong divisor count.
\end{trajblock}
\raggedright\textit{\textcolor{black!55}{Example model behavior: GPT-5.5}}
\end{failurecase}
}

\vspace{3pt}
\caption{
\textbf{Representative state-maintenance and intermediate-reasoning failures in \textsc{K-BrowseComp}.}
Each panel shows the question, gold answer, required trajectory state, intended gold trajectory, model search queries, and the point where the trajectory first diverges.
F7 illustrates a constraint-tracking failure, where the model commits to a locally plausible K-pop group candidate without enforcing the full intersection of parallel constraints.
F8 illustrates an intermediate reasoning failure, where the model successfully retrieves relevant intermediate quantities but produces an incorrect final divisor count during numerical reasoning.
}
\label{fig:failure_modes_appendix_f7_f8}
\end{figure*}


\begin{figure*}[t]
\centering
\scriptsize
\setlength{\fboxsep}{0pt}

\begin{minipage}[t]{0.492\textwidth}
\begin{tcolorbox}[
  enhanced,
  colback=white,
  colframe=black!65,
  boxrule=0.55pt,
  arc=2pt,
  left=4pt,
  right=4pt,
  top=4pt,
  bottom=4pt
]
\centering
\textbf{(a) Candidate-ledger failure}
\vspace{2pt}

\begin{trajblock}{kbBlue}{\kbBadge{kbBlue}{Q}\hspace{3pt} Question}
골든디스크어워즈 디지털 음원부분에서 3곡 이상으로 수상을 한 Kpop 그룹으로,
발매한 노래의 가사 중 멤버의 이름이 들어간 노래가 있고, 멤버 중 한명이
고등학교를 배경으로 한 드라마에 출연한 적이 있으며, 멤버들의 국적이
2개 이상으로 이루어진 그룹은?

\eng{Which K-pop group satisfies all of the following conditions: the group won awards for at least three songs in the Digital Song Division of the Golden Disc Awards, has a released song whose lyrics contain the name of one of the members, has at least one member who appeared in a drama set in a high school, and consists of members from at least two different nationalities?}

\vspace{2pt}
\textcolor{kbBlue!80!black}{\textbf{Gold:}} 미쓰에이 (\textit{miss A})
\end{trajblock}

\begin{trajblock}{black}{\kbBadge{kbGray}{S}\hspace{3pt} Required state}
Maintain one candidate ledger across all parallel constraints.
\end{trajblock}

\begin{trajblock}{kbGreen}{\kbBadge{kbGreen}{G}\hspace{3pt} Gold state}
\begin{itemize}[leftmargin=1.25em,itemsep=0pt,topsep=0pt]
    \item Golden Disc digital-song winners with $\geq 3$ awarded songs
    \item[$\cap$] lyrics mentioning a member name
    \item[$\cap$] member appeared in a high-school drama
    \item[$\cap$] multinational membership
\end{itemize}
\vspace{1pt}
\textcolor{kbGreen!75!black}{$\rightarrow$ \textbf{미쓰에이(miss A)}}
\end{trajblock}

\begin{trajblock}{kbRed}{\kbBadge{kbRed}{M}\hspace{3pt} Model trajectory}
\textit{Search queries}
\searchquery{골든디스크어워즈 디지털 음원 부문 수상 3곡 이상 Kpop 그룹 
\eng{Golden Disc Awards most digital song bonsang wins Kpop group 3 or more}}

\vspace{1pt}
\rstep{1} Search broad award-history candidates.\\
\rstep{2} Drift toward high-visibility artists such as IU / BTS.\\
\rstep{3} Discuss lyric, drama, and nationality clues separately.\\
\rstep{4} Never merge them into one shared candidate table.\\
\rstep{5} \textcolor{kbRed!80!black}{\textbf{Result: I don't know}}
\end{trajblock}

\begin{trajblock}{kbOrange}{\kbBadge{kbOrange}{!}\hspace{3pt} Why this fails}
\begin{itemize}[leftmargin=1.1em,itemsep=-2pt,topsep=0pt]
    \item Searches relevant award evidence, but never converts it into a shared candidate set.
    \item Treats the remaining clues as separate unresolved checks.
    \item Fails to intersect all constraints on the same group.
\end{itemize}
\end{trajblock}

\vspace{1pt}
\raggedright\textit{\textcolor{black!55}{Example model behavior: K-EXAONE-236B-A23B}}
\end{tcolorbox}
\end{minipage}
\hfill
\begin{minipage}[t]{0.492\textwidth}
\begin{tcolorbox}[
  enhanced,
  colback=white,
  colframe=black!65,
  boxrule=0.55pt,
  arc=2pt,
  left=4pt,
  right=4pt,
  top=4pt,
  bottom=4pt
]
\centering
\textbf{(b) Answer-side candidate capture}
\vspace{2pt}

\begin{trajblock}{kbBlue}{\kbBadge{kbBlue}{Q}\hspace{3pt} Question}
2026년 4월 기준으로 기초 화장품과, 색조 화장품, 남성용 화장품과,
여성 케어 총 네가지 브랜드를 소유중인 한국 회사의 브랜드 중
2005년 이전에 설립된 헬스 앤 뷰티 스토어에 2018년 이전에 입점한
브랜드에서 2024--2025년에 해당 스토어에서 수상한 제품의 이름은?

\eng{As of April 2026, among the brands owned by a Korean company that operates four categories of brands, namely basic skincare, color cosmetics, men's cosmetics, and feminine care, what is the name of the product that won an award at that health-and-beauty store in 2024--2025, from a brand that had entered a health-and-beauty store established before 2005 prior to 2018?}

\vspace{2pt}
\textcolor{kbBlue!80!black}{\textbf{Gold:}} 어성초 흔적 에센스 패드 (Abib Heartleaf Spot Pad Calming Touch)
\end{trajblock}

\begin{trajblock}{black}{\kbBadge{kbGray}{S}\hspace{3pt} Required state}
Preserve the chain:
company $\rightarrow$ brand $\rightarrow$ H\&B store $\rightarrow$ award product.
\end{trajblock}

\begin{trajblock}{kbGreen}{\kbBadge{kbGreen}{G}\hspace{3pt} Gold trajectory}
\gstep{1} Identify the Korean company satisfying the brand-category constraint.\\
\gstep{2} Fix the qualifying brand and H\&B store.\\
\gstep{3} Resolve the store as Olive Young.\\
\gstep{4} Search 2024--2025 award results for that fixed brand.\\
\gstep{5} \textcolor{kbGreen!75!black}{\textbf{Result: 어성초 흔적 에센스 패드} (Abib Heartleaf Spot Pad Calming Touch)}
\end{trajblock}

\begin{trajblock}{kbRed}{\kbBadge{kbRed}{M}\hspace{3pt} Model trajectory}
\textit{Raw search queries}
\searchquery{2024 2025 헬스 앤 뷰티 스토어(H\&S) 어워드 수상 제품 리스트 기초 색조 남성 여성케어 2005년 이전 설립 브랜드
\eng{2024--2025 H\&B store award winners skincare makeup men's care feminine care}}
\searchquery{헬스 앤 뷰티 스토어 H\&S 화장품 입점 브랜드 2018년 이전
\eng{H\&B store cosmetic brands before 2018}}
\searchquery{올리브영 2024 2025 어워드 수상 제품 브랜드 2005년 설립 이전
\eng{Olive Young 2024--2025 award winners pre-2005 brands}}

\vspace{1pt}
\rstep{1} Search Olive Young / H\&B award products early.\\
\rstep{2} Search store-entry evidence before fixing the upstream brand.\\
\rstep{3} Anchor on the CLIO / Goodal / Peripera branch.\\
\rstep{4} Collect locally plausible award products from that branch.\\
\rstep{5} \textcolor{kbRed!80!black}{\textbf{Result: multiple CLIO / Goodal / Peripera products}}
\end{trajblock}

\begin{trajblock}{kbOrange}{\kbBadge{kbOrange}{!}\hspace{3pt} Why this fails}
\begin{itemize}[leftmargin=1.1em,itemsep=-2pt,topsep=0pt]
    \item Searches answer-side award evidence before fixing the upstream brand identity.
    \item Commits to a locally plausible CLIO/Goodal/Peripera branch.
    \item Returns multiple locally supported products instead of the unique gold answer.
\end{itemize}
\end{trajblock}

\vspace{1pt}
\raggedright\textit{\textcolor{black!55}{Example model behavior: Qwen3.6-35B-A3B}}
\end{tcolorbox}
\end{minipage}

\vspace{3pt}
\caption{
\textbf{Representative trajectory-level failures in \textsc{K-BrowseComp}.}
Each panel shows the question, the required intermediate state, the gold state or trajectory, and the model's raw search queries.
Panel~(a) shows a candidate-ledger failure: the model retrieves partially relevant award evidence, but does not merge all constraints into a shared candidate set, failing to identify 미쓰에이(\textit{miss A}).
Panel~(b) shows answer-side candidate capture: the model searches the award side of the question before fixing the upstream company and brand, then anchors on a locally plausible branch and returns multiple products instead of the gold answer, 어성초 흔적 에센스 패드 (\textit{Abib Heartleaf Spot Pad Calming Touch}).
}
\label{fig:trajectory_cases_appendix}
\end{figure*}


\begin{figure*}[t]
\centering
\scriptsize
\setlength{\fboxsep}{0pt}

\begin{minipage}[t]{0.985\textwidth}
\begin{tcolorbox}[
  enhanced,
  colback=white,
  colframe=black!65,
  boxrule=0.55pt,
  arc=2pt,
  left=5pt,
  right=5pt,
  top=5pt,
  bottom=5pt
]
\centering
\textbf{Shallow evidence control after a broad search}
\vspace{2pt}

\begin{trajblock}{kbBlue}{\kbBadge{kbBlue}{Q}\hspace{3pt} Question}
형제가 새 왕조 개창 후 몇 년이 지나지 않아 문과 급제자 명단에 이름을 올렸으며,
동성동본의 현대 운동선수 가운데 라리가와 리그 1 무대를 모두 경험한 사람이 있고,
감찰과 관련된 관직을 역임한 적이 있으며, 사후 그의 고장에 그를 기리는 서원이 세워진 인물은?

\eng{Which person satisfies the following conditions: his brothers appeared on the civil-service examination list shortly after the founding of a new dynasty; a modern athlete of the same surname and clan experienced both La Liga and Ligue~1; he held an inspection-related office; and a memorial academy was established in his hometown after his death?}

\vspace{1pt}
\textcolor{kbBlue!80!black}{\textbf{Gold:}} 이명성 \eng{Lee Myung Seong}
\end{trajblock}

\begin{trajblock}{kbState}{\kbBadge{kbState}{S}\hspace{3pt} Required state}
Maintain one historical-person candidate ledger across exam records, surname/clan evidence, the modern-athlete clue, office history, and memorial-site evidence.
\end{trajblock}

\begin{trajblock}{kbGreen}{\kbBadge{kbGreen}{G}\hspace{3pt} Gold trajectory}
\gstep{1} Resolve the same-surname-and-clan clue through the modern athlete who played in both La Liga and Ligue~1.\\
\gstep{2} Use that surname/clan state to search historical figures with brothers listed in early civil-service examination records.\\
\gstep{3} Filter candidates by inspection-related office history.\\
\gstep{4} Verify that a memorial academy was established in the candidate's hometown.\\
\gstep{5} Return the remaining candidate.\\
\textcolor{kbGreen!75!black}{$\rightarrow$ \textbf{이명성} \eng{Lee Myung Seong}}
\end{trajblock}

\begin{trajblock}{kbRed}{\kbBadge{kbRed}{M}\hspace{3pt} Model trajectory}

\textit{Raw search query}
\searchquery{형제 새 왕조 개창 후 문과 급제자, 동성동본 현대 운동선수 라리가와 리그 1 경험, 감찰 관련 관직, 사후 서원 설립 인물

\eng{civil-service passer after brothers founded a new dynasty same-clan modern athlete La Liga and Ligue 1 experience censorship-related office posthumous academy figure}
}

\vspace{1pt}
\rstep{1} The model begins with one broad query that mixes all constraints into a single search.\\
\rstep{2} The search returns partially relevant historical material, including exam-list pages and visible names such as 정황, 민영환, and 김종서.\\
\rstep{3} The model summarizes these candidates one by one, checking some local clues such as examination records, inspection-related offices, or memorial evidence.\\
\rstep{4} The same-origin modern-athlete clue is not used as a binding constraint on the candidate set.\\
\rstep{5} The trajectory never forms a shared ledger of remaining candidates, so evidence about different people is compared only locally.\\
\rstep{6} \textcolor{kbRed!80!black}{\textbf{Result: I don't know}}
\end{trajblock}

\begin{trajblock}{kbOrange}{\kbBadge{kbOrange}{!}\hspace{3pt} Failure point}
The model reaches relevant Korean historical evidence, but treats retrieved snippets as candidate descriptions instead of filters over one shared candidate state.
The failure occurs after retrieval: plausible names are discussed, but the full constraint set is never enforced on the same person.
\end{trajblock}

\vspace{1pt}
\raggedright\textit{\textcolor{black!55}{Example model behavior: A.X-4.0}}
\end{tcolorbox}
\end{minipage}

\vspace{5pt}
\caption{
\textbf{Shallow evidence-control failure in \texttt{A.X-4.0}.}
The example shows a Korean historical-entity question that requires maintaining one candidate ledger across civil-service examination records, surname/clan evidence, a modern-athlete clue, office history, and memorial-site evidence.
The model issues a single broad query that concatenates most constraints, retrieves partially relevant historical candidates, and discusses locally plausible names such as 정황, 민영환, and 김종서.
However, it does not enforce all constraints on the same candidate state, leaving the same-origin modern-athlete clue unbound and ending with \textit{I don't know}.
This illustrates weak evidence control after retrieval: relevant snippets are found, but they are not converted into filters over a stable candidate ledger.
}
\label{fig:ax_weak_evidence_control}
\end{figure*}


\begin{figure*}[t]
\centering
\scriptsize
\setlength{\fboxsep}{0pt}
\begin{minipage}[t]{0.985\textwidth}
\begin{tcolorbox}[
  enhanced,
  colback=white,
  colframe=black!65,
  boxrule=0.55pt,
  arc=2pt,
  left=5pt,
  right=5pt,
  top=5pt,
  bottom=5pt
]
\centering
\textbf{Cross-source chain drift after successful initial retrieval}
\vspace{2pt}

\begin{trajblock}{kbBlue}{\kbBadge{kbBlue}{Q}\hspace{3pt} Question}
2025년 7월 3째주에 카카오톡 무료 이모티콘 이벤트에서 배포한 이모티콘 중
더 작은 동물과 관련된 캐릭터 이모티콘의 제작자의 유튜브 채널 가입일은?

\eng{Among the emoticons distributed in the KakaoTalk free-emoticon event in the third week of July 2025, what is the YouTube channel join date of the creator of the character emoticon related to the smaller animal?}

\vspace{1pt}
\textcolor{kbBlue!80!black}{\textbf{Gold:}} 2021년 8월 9일
\end{trajblock}

\begin{trajblock}{kbState}{\kbBadge{kbState}{S}\hspace{3pt} Required state}
Preserve the dependency chain:
event week $\rightarrow$ distributed emoticons $\rightarrow$ smaller-animal character $\rightarrow$ creator $\rightarrow$ official YouTube channel metadata.
\end{trajblock}

\begin{trajblock}{kbGreen}{\kbBadge{kbGreen}{G}\hspace{3pt} Gold trajectory}
\gstep{1} Identify the KakaoTalk free-emoticon event for the third week of July 2025.\\
\gstep{2} List the animal-related character emoticons distributed in that event.\\
\gstep{3} Determine which distributed character corresponds to the smaller animal.\\
\gstep{4} Find the creator of that fixed character.\\
\gstep{5} Open the creator's official YouTube channel metadata and read the channel join date.\\
\textcolor{kbGreen!75!black}{$\rightarrow$ \textbf{2021년 8월 9일}}
\end{trajblock}

\begin{trajblock}{kbRed}{\kbBadge{kbRed}{M}\hspace{3pt} Model trajectory}

\textit{Raw search queries}
\searchquery{카카오톡 무료 이모티콘 이벤트 2025년 7월 셋째주 작은 동물 캐릭터}
\searchquery{극철이 망그래진 헤터 오둑이 연락켓 YouTube 참니달콬라 가입일}
\searchquery{극락 쿼커 이모티콘 제작자 YouTube}
\searchquery{"긍" "긍" "긍" "ompp" YouTube}
\searchquery{긍비금비 카카오 이모티콘 작가 극락 쿼커}

\vspace{1pt}
\rstep{1} The first search enters the right event region and retrieves the Kakao announcement and related reports.\\
\rstep{2} The retrieved evidence lists candidate emoticons such as 곽철이, 망그러진 햄터, 오둥이, 극락 쿼카, and 포테토뭉.\\
\rstep{3} The model does not freeze the event-side candidate state before moving to creator search.\\
\rstep{4} Follow-up queries \textbf{begin to mix misspelled character names, malformed creator strings}, and unrelated YouTube terms.\\
\rstep{5} The model finds a plausible creator-like entity, 긍비금비, and a YouTube channel date, but cannot verify that this creator is the maker of the target smaller-animal emoticon.\\
\rstep{6} The final explanation reports this unresolved link and returns \textcolor{kbRed!80!black}{\textbf{I don't know}}.
\end{trajblock}

\begin{trajblock}{kbOrange}{\kbBadge{kbOrange}{!}\hspace{3pt} Failure point}
The initial search retrieves the relevant Kakao event evidence, but the model fails to stabilize the intermediate target emoticon before moving to creator search.
After this point, the follow-up queries contain corrupted character names and no longer preserve the event--emoticon--creator--channel dependency chain.
\end{trajblock}

\vspace{1pt}
\raggedright\textit{\textcolor{black!55}{Example model behavior: K-EXAONE-236B-A23B}}
\end{tcolorbox}
\end{minipage}

\vspace{3pt}
\caption{
\textbf{Cross-source chain drift in \texttt{K-EXAONE-236B-A23B}.}
The example shows a KakaoTalk free-emoticon question whose answer requires preserving the dependency chain from event week to distributed emoticons, smaller-animal character, creator, and official YouTube channel metadata.
The model's first query reaches the relevant Kakao event evidence and retrieves the candidate emoticon list, including 곽철이, 망그러진 햄터, 오둥이, 극락 쿼카, and 포테토뭉.
After this initial retrieval, the model fails to stabilize the target emoticon before moving to creator search.
Its follow-up queries contain corrupted character names and malformed creator/channel strings, and the trajectory ends without a verified creator--channel link.
This illustrates weak evidence control across sources: the model reaches the right web region, but loses the intermediate entity chain needed to license the final answer.
}
\label{fig:exaone_cross_source_chain_drift}
\end{figure*}



\begin{figure*}[t]
\centering
\scriptsize
\setlength{\fboxsep}{0pt}

\begin{minipage}[t]{0.985\textwidth}
\begin{tcolorbox}[
  enhanced,
  colback=white,
  colframe=black!65,
  boxrule=0.55pt,
  arc=2pt,
  left=5pt,
  right=5pt,
  top=5pt,
  bottom=5pt
]
\centering
\textbf{Metadata-field extraction failure after reaching the correct repository}
\vspace{2pt}

\begin{trajblock}{kbBlue}{\kbBadge{kbBlue}{Q}\hspace{3pt} Question}
극지연구소는 2015년 2월, 쇄빙연구선 아라온호를 활용한 남극 중앙해령 탐사를 통해
현무암질 해양 지각에 빙하기-간빙기 순환의 기록이 남아 있다는 사실을 세계 최초로 규명했고,
그 성과가 국제 학술지 `사이언스(Science)'에 논문으로 실렸다는 내용을 알리는 보도자료를 냈다.
이 보도자료 원문은 극지연구소가 운영하는 기관 리포지터리에 PDF 파일 한 건으로 등록되어 일반에 공개되어 있다.
해당 자료 상세 페이지에서 이 첨부 PDF 파일에 대해 표시하고 있는 파일 용량은 무엇인가?

\eng{In February 2015, the Korea Polar Research Institute announced that research using the icebreaker Araon had found evidence of glacial-interglacial cycles in basaltic oceanic crust at the Antarctic mid-ocean ridge, with the result published in \textit{Science}.
The original press release is publicly registered as a PDF file in the institute repository.
What file size is displayed for the attached PDF on the repository detail page?}

\vspace{1pt}
\textcolor{kbBlue!80!black}{\textbf{Gold:}} 698.85 kB
\end{trajblock}

\begin{trajblock}{kbState}{\kbBadge{kbState}{S}\hspace{3pt} Required state}
Preserve the target repository item and extract the file-size field from the attached PDF metadata, not from a nearby PDF or another repository record.
\end{trajblock}

\begin{trajblock}{kbGreen}{\kbBadge{kbGreen}{G}\hspace{3pt} Gold trajectory}
\gstep{1} Identify the 2015 KOPRI press release about the Araon Antarctic mid-ocean ridge study.\\
\gstep{2} Open the KOPRI repository detail page for the target press release.\\
\gstep{3} Locate the attached PDF in the item metadata.\\
\gstep{4} Read the displayed file-size value exactly as written on the page.\\
\textcolor{kbGreen!75!black}{$\rightarrow$ \textbf{698.85 kB}}
\end{trajblock}

\begin{trajblock}{kbRed}{\kbBadge{kbRed}{M}\hspace{3pt} Model trajectory}

\textit{Raw search queries}
\searchquery{극지연구소 2015년 2월 아라온호 남극 중앙해령 사이언스 보도자료

site:repository.kopri.re.kr "6-1.pdf" 4533

\eng{Korea Polar Research Institute February 2015 Araon Antarctic mid-ocean ridge science press release}
}

\vspace{1pt}
\rstep{1} The model retrieves the correct KOPRI topic and reaches the repository source family.\\
\rstep{2} The search results include the target repository item and several nearby PDF records with similar filenames.\\
\rstep{3} The model does not preserve the target item when reading the file metadata.\\
\rstep{4} It selects a nearby incorrect file-size value.\\
\rstep{5} \textcolor{kbRed!80!black}{\textbf{Result:}} 1.56 MB
\end{trajblock}

\begin{trajblock}{kbOrange}{\kbBadge{kbOrange}{!}\hspace{3pt} Failure point}
The failure occurs after relevant retrieval.
The model reaches the correct repository neighborhood, but does not extract the exact metadata field from the target item.
This is a semi-structured parsing failure (F4).
\end{trajblock}

\vspace{1pt}
\raggedright\textit{\textcolor{black!55}{Example model behavior: Gemini-3.1-Flash-Lite}}
\end{tcolorbox}
\end{minipage}

\vspace{5pt}
\caption{
\textbf{Semi-structured metadata parsing failure in the \textsc{Synthetic} split.}
The question asks for the file size displayed for an attached PDF on a KOPRI repository detail page.
The model reaches the correct source neighborhood and retrieves the relevant press-release context, but it does not preserve the target repository item when reading the file metadata.
It returns a nearby incorrect file-size value, 1.56 MB, instead of the gold value, 698.85 kB.
}
\label{fig:synthetic_filesize_failure}
\end{figure*}


\begin{figure*}[t]
\centering
\scriptsize
\setlength{\fboxsep}{0pt}

\begin{minipage}[t]{0.985\textwidth}
\begin{tcolorbox}[
  enhanced,
  colback=white,
  colframe=black!65,
  boxrule=0.55pt,
  arc=2pt,
  left=5pt,
  right=5pt,
  top=5pt,
  bottom=5pt
]
\centering
\textbf{Game-level constraint tracking failure after identifying the target player}
\vspace{2pt}

\begin{trajblock}{kbBlue}{\kbBadge{kbBlue}{Q}\hspace{3pt} Question}
2022 정규시즌에 평균자책점, 탈삼진, 이닝 세 부문에서 모두 리그 1위에 오른 키움 히어로즈 소속의 한 우완 투수가 있다.
이 선수가 2026 정규시즌에 등판한 경기들 가운데, 그 경기에서 상대 타자들에게 기록당한 피안타율이 가장 높았던 경기는 단 한 경기뿐이다.
그 경기의 날짜는 언제인가?
답은 기록 페이지의 표기 방식인 `MM.DD' 형식으로 적어라.

\eng{A right-handed pitcher for the Kiwoom Heroes led the league in ERA, strikeouts, and innings pitched during the 2022 regular season.
Among the games in which this player appeared during the 2026 regular season, exactly one game had the highest opponent batting average against him.
What was the date of that game?
Answer in the record-page format, `MM.DD'.}

\vspace{1pt}
\textcolor{kbBlue!80!black}{\textbf{Gold:}} 05.08
\end{trajblock}

\begin{trajblock}{kbState}{\kbBadge{kbState}{S}\hspace{3pt} Required state}
Maintain one game-level ledger for the target pitcher's 2026 appearances, compare opponent AVG values across games, and select the unique maximum.
\end{trajblock}

\begin{trajblock}{kbGreen}{\kbBadge{kbGreen}{G}\hspace{3pt} Gold trajectory}
\gstep{1} Identify the 2022 Kiwoom pitcher who led ERA, strikeouts, and innings.\\
\gstep{2} Fix the target player as 안우진 \textit{An Woo-jin}.\\
\gstep{3} Retrieve the player's 2026 game-level pitching records.\\
\gstep{4} Compare opponent AVG values for each appearance.\\
\gstep{5} Select the single game with the highest opponent AVG and return its date.\\
\textcolor{kbGreen!75!black}{$\rightarrow$ \textbf{05.08}}
\end{trajblock}

\begin{trajblock}{kbRed}{\kbBadge{kbRed}{M}\hspace{3pt} Model trajectory}

\textit{Raw search queries}
\searchquery{2022 KBO 평균자책점 탈삼진 이닝 1위 키움 히어로즈 투수

\eng{KBO pitcher from Kiwoom Heroes who led the league in ERA, strikeouts, and innings pitched in 2022}
}

\searchquery{안우진 2026 시즌 등판 기록

\eng{Ahn Woo-jin 2026 season appearance record}
}

\searchquery{안우진 2026 시즌 등판 경기 날짜 피안타율

\eng{Ahn Woo-jin 2026 season game dates opponent batting average}
}

\searchquery{안우진 "피안타율이 가장 높았던 경기는 단 한 경기뿐" 2026

\eng{Ahn Woo-jin "only one game had the highest opponent batting average" 2026}
}

\vspace{1pt}
\rstep{1} The model correctly identifies the target player as 안우진 \eng{An Woo-jin}.\\
\rstep{2} It then searches for 2026 appearance records and game-level opponent AVG evidence.\\
\rstep{3} The searches reach relevant baseball-record pages and news reports, including KBO player-record pages and reports about specific 2026 appearances.\\
\rstep{4} However, the trajectory does not maintain a complete game-level ledger with date, appearance, and opponent AVG fields.\\
\rstep{5} The model selects a plausible later date from the retrieved appearance evidence without verifying the unique maximum AVG condition.\\
\rstep{6} \textcolor{kbRed!80!black}{\textbf{Result:}} 05.14
\end{trajblock}

\begin{trajblock}{kbOrange}{\kbBadge{kbOrange}{!}\hspace{3pt} Failure point}
The target player is correctly recovered, and the later searches enter the relevant 2026 record space.
The failure occurs at the constraint-maintenance step: the model does not build a stable game-level table and therefore cannot compare the opponent AVG field across all appearances.
\end{trajblock}

\vspace{1pt}
\raggedright\textit{\textcolor{black!55}{Example model behavior: Gemini-3.1-Flash-Lite}}
\end{tcolorbox}
\end{minipage}

\vspace{5pt}
\caption{
\textbf{Constraint-tracking failure in a \textsc{Synthetic} KBO record question.}
The question identifies 안우진 \eng{An Woo-jin} from the 2022 league-leading pitching clues and requires comparing opponent AVG values across his 2026 game-level records.
The model recovers the target player and enters the relevant baseball-record region, but does not keep a stable game-level ledger for the final comparison.
It selects 05.14, while the gold date is 05.08.
This example shows how the \textsc{Synthetic} split stresses constraint maintenance and exact selection after the main entity has already been found.
}
\label{fig:synthetic_kbo_constraint_failure}
\end{figure*}

\end{document}